\documentclass[journal,twoside,web]{ieeecolor}
\usepackage{generic}
\usepackage{amsmath,amssymb,amsfonts}
\usepackage{algorithmic}
\usepackage{graphicx}
\usepackage{algorithm,algorithmic}
\usepackage{textcomp}

\usepackage[style=ieee]{biblatex}
\bibliography{refs}

\usepackage{hyperref}

\def\BibTeX{{\rm B\kern-.05em{\sc i\kern-.025em b}\kern-.08em
    T\kern-.1667em\lower.7ex\hbox{E}\kern-.125emX}}
\usepackage{subfigure}
\usepackage{caption}
\usepackage{multirow}
\usepackage{url}
\usepackage{booktabs}
\usepackage{nicefrac}
\usepackage{xcolor}
\begin{document}
% \title{CARNA: Characterizing Advanced Heart Failure Risk and HemodyNAmic Phenotypes using Learned Multi-Valued Decision Diagrams}
\title{CARNA: Characterizing Advanced heart failure Risk and hemodyNAmic phenotypes using learned multi-valued decision diagrams}
\author{
% First A. Author, \IEEEmembership{Fellow, IEEE}, Second B. Author, and Third C. Author Jr., \IEEEmembership{Member, IEEE}
Josephine Lamp, Yuxin Wu, Steven Lamp, Prince Afriyie, Kenneth Bilchick, Lu Feng, Sula Mazimba
% Josephine Lamp, Yuxin Wu, Steven Lamp, Prince Afriyie, Nicholas Ashur, Kenneth Bilchick, Khadijah Breathett, Younghoon Kwon, Song Li, Nishaki Mehta, Edward Rojas Pena, Lu Feng, Sula Mazimba
\thanks{
This material is based upon work supported by the National Science Foundation Graduate Research Fellowship under Grant No. (1842490). This work was based on one of the winning solutions to the NHLBI Big Data Analysis Challenge: Creating New Paradigms for Heart Failure Research. 
% Dr. Breathett received support from National Heart, Lung, and Blood Institute R56HL159216, K01HL142848, and L30HL148881. 
% Any opinion, findings, and conclusions or recommendations expressed in this material are those of the authors(s) and do not necessarily reflect the views of the grant sponsors.
}
\thanks{Josephine Lamp, Steven Lamp and Lu Feng are with the Department of Computer Science, University of Virginia, Charlottesville, VA, USA (e-mail: \{jl4rj; vxn3kr; lu.feng\}@virginia.edu)}
\thanks{Yuxin Wu is with the Department of Computer Science, University of California, Los Angeles, CA, USA (e-mail: yuxinwu98611@g.ucla.edu)}
\thanks{Prince Afriyie is with the Department of Statistics, University of Virginia, Charlottesville, VA, USA (e-mail: pa5jg@virginia.edu)}
\thanks{Kenneth Bilchick and Sula Mazimba are with the Department of Cardiovascular Medicine, University of Virginia, Charlottesville, VA, USA (e-mail: \{KCB7F; SM8SD\}@uvahealth.org)}
% \thanks{The next few paragraphs should contain 
% the authors' current affiliations, including current address and e-mail. For 
% example, First A. Author is with the National Institute of Standards and 
% Technology, Boulder, CO 80305 USA (e-mail: author@boulder.nist.gov). }
}

\maketitle

% The abstract must be between 150--250 words. 
\begin{abstract}
% Heart failure (HF) is a complex disease condition with high morbidity and mortality. 
Early identification of high risk heart failure (HF) patients is key to timely allocation of life-saving therapies. Hemodynamic assessments can facilitate risk stratification and enhance understanding of HF trajectories. However, risk assessment for HF is a complex, multi-faceted decision-making process that can be challenging. 
Previous risk models for HF do not integrate invasive hemodynamics or support missing data, and use statistical methods prone to bias or machine learning methods that are not interpretable. 
To address these limitations, this paper presents CARNA (Characterizing Advanced heart failure
Risk and hemodyNAmic phenotypes), a hemodynamic risk stratification and phenotyping framework for advanced HF that takes advantage of the explainability and expressivity of machine learned Multi-Valued Decision Diagrams (MVDDs). This interpretable framework learns risk scores that predict the probability of patient outcomes, and outputs descriptive patient phenotypes (sets of features and thresholds) that characterize each predicted risk score. Specifically, we develop an innovative methodology to first learn a risk stratification using hierarchical clustering, and develop a training regime to learn MVDDs that predict risk scores and output patient phenotypes. CARNA incorporates invasive hemodynamics and can make predictions on missing data. The CARNA models were trained and validated using a total of five advanced HF patient cohorts collected from previous trials, and compared with six established HF risk scores and three traditional ML risk models. CARNA provides robust risk stratification, outperforming all previous benchmarks. 
Although focused on advanced HF, the CARNA framework is general purpose and can be used to learn risk stratifications for other diseases and medical applications. Moreover, to facilitate practical use, we provide an extensible, open source tool implementation.
% with averaged AUCs of 0.938 to 0.965 and 0.871 to 0.996 for the Invasive Hemodynamics and All Features scores, respectively.
\end{abstract}


%Note: 12 page limit, including refs
\section{Introduction}\label{sec:introduction}
\IEEEPARstart{H}{eart} failure (HF) is a complex disease condition with high morbidity and mortality~\cite{dunlay2021advanced}. On a fundamental level, HF is defined by the inability of the heart to deliver adequate blood flow to the body without an elevation in cardiac filling pressures~\cite{verbrugge2020altered}. 
Identifying high risk advanced HF patients early on in the care continuum is critical for timely allocation of advanced, life-saving therapies such as mechanical support, device implantation or transplant allocation. 
% For example, HF admission is a key event that is associated with increased mortality~\cite{ahmed2006higher}. 
% Determining patient risk states in the context of outcomes such as mortality and rehospitalization is useful, as additional rehospitalizations are associated with increased mortality rates~\cite{ahmed2006higher}. 
Due to high variability in patient conditions and complexity of the disease, determining patient risk involves a challenging, multi-faceted decision making process that places a high burden on clinicians~\cite{allen2012decision}. 
% Risk assessment for HF is a complex, multi-faceted decision-making process that can be challenging~\cite{allen2012decision}.
Hemodynamic assessments can facilitate risk stratification and enhance understanding of HF trajectories~\cite{bilchick2018clinical}.
Hemodynamics provide measures of cardiovascular function, and quantify distributions of pressures and flows within the heart and circulatory system~\cite{hsu2022hemodynamics}. %secomb2016hemodynamics 
However, obtaining a comprehensive picture of the patient state from these, particularly in the context of treatment-guiding outcomes, is difficult~\cite{borlaug2011invasive}. 
% How handle potentially conflicting patient variability in presentation of hf condition states and outcomes and innate patient variability

% % issues with previous methods
% Established HF risk scores such as the Seattle Heart Failure Risk model~\cite{levy2006seattle} do not incorporate invasive hemodynamics and may suffer from limitations associated with statistical and naive 
% % machine learning 
% models, including being prone to bias and/or lacking mechanisms to handle missing data~\cite{di2020evaluating,canepa2018performance,codina2021head}. 
% %ml do better but opportunities of ml not been realized
% When carefully constructed (e.g., to reduce bias), machine learning (ML) models present a promising opportunity to outperform traditional 
% % statistical 
% risk assessment methods, especially when dealing with large, high dimensional data~\cite{greenberg2021machine}. However, despite the promise of machine learning for HF risk stratification, 
% Despite promise of ML in HF and risk stratification, the opportunities of their use have not been realized~\cite{mpanya2021predicting}.because of issues with x y and z

Many established HF risk scores such as the Seattle Heart Failure Risk model~\cite{levy2006seattle} use statistical or naive models which are difficult to optimize and may be prone to bias~\cite{di2020evaluating,canepa2018performance,codina2021head}. 
% When carefully constructed (e.g., to reduce bias), 
Machine learning (ML) models present a promising opportunity to outperform traditional risk assessment methods, especially when dealing with large, high-dimensional data~\cite{greenberg2021machine}. However, despite the promise of machine learning for HF risk stratification, ML-based risk scores remain unpopular due to modest model performance and issues with model interpretability~\cite{mpanya2021predicting}. 
Moreover, no previous models (statistical of ML-based) incorporate invasive hemodynamics, or contain mechanisms to handle missing data.

\begin{figure}[t]
    \centering
    \includegraphics[width=\linewidth]{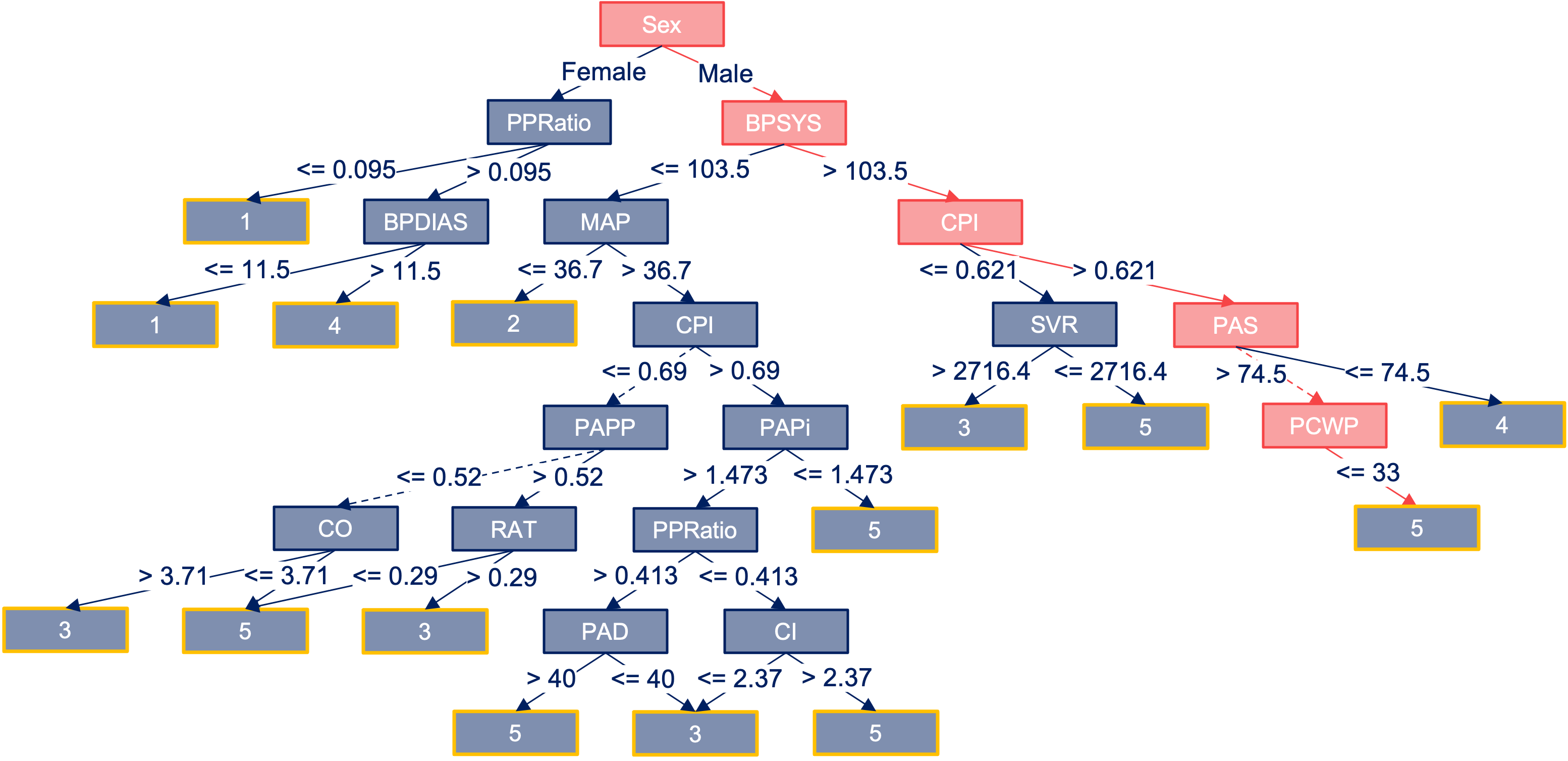}
    \caption{Example MVDD for the Invasive Hemodynamic Feature Set and DeLvTx Outcome. Dotted lines represent “or” boolean operators, and solid lines represent “and” boolean operators. The leaf nodes highlighted in yellow indicate the risk score. The highlighted red path indicates the example phenotype of Sex = Male $\land$ BPSYS $>$ 103.5 $\land$ CPI $>$ 0.621 $\land$ (PAS $>$ 74.5 $\lor$ PCWP $\leq$ 33) = Score 5. }
    % BPDIAS indicates blood pressure diastolic; BPSYS, blood pressure systolic; CI, cardiac index; CO, cardiac output; CPI, cardiac power index; DeLvTx, composite endpoint of death, left ventricular assist device (LVAD) implantation or transplantation; MAP, mean arterial pressure; PAD, Pulmonary Artery Pressure Diastolic; PAPi, Pulmonary Artery Pulsatility Index; PAPP, Pulmonary Arterial Proportional Pulse Pressure (Right Side); PAS, Pulmonary Artery Pressure Systolic; PCWP, Pulmonary Capillary Wedge Pressure; PPRatio, Ratio of systemic pulse pressure / heart rate; RAT, Ratio of right atrial pressure (RAP) to Pulmonary Capillary Wedge Pressure (PCWP); SVR, systemic vascular resistance.}
    \label{fig:example-mvdd}
\end{figure}

To address these limitations, this paper develops and validates an advanced HF hemodynamic risk stratification framework entitled CARNA (Characterizing Advanced heart failure Risk and hemodyNAmic phenotypes)\footnote{So named for the roman healing goddess who presides over the heart.}. We harness the explainability and expressivity of machine learned Multi-Valued Decision Diagrams (MVDDs) to learn a risk score that predicts the probability of patient outcomes, including mortality and rehospitalization, and provide descriptive patient phenotypes. 
MVDDs are discrete structures representing logical functions in directed, acyclic graphs where nodes represent features, edges represent logical operators (“and”, “or”) with parameter threshold values, and leaf nodes represent the final score classification~\cite{srinivasan1990algorithms}. An example MVDD is shown in Figure~\ref{fig:example-mvdd}. Due to their use of logical operators, MVDDs can handle missing data, as multiple substitutable features may contribute to the same score prediction. Moreover, the ``path" through the MVDD may be returned to provide a descriptive patient phenotype that characterizes the score. MVDDs have typically been applied in optimization and model checking contexts~\cite{bergman2016decision}, and they do not inherently learn a risk stratification. Therefore, we develop an innovative methodology within our framework to first learn a risk stratification using a hierarchical clustering algorithm, and then develop a training regime to train the MVDDs on the learned risk scores and output explainable phenotypes. Although focused on advanced HF, CARNA is a general purpose risk stratification and phenotyping framework that can be used for other diseases and medical applications.

In summary, we present the following contributions:
\begin{enumerate}
\item We develop CARNA, an interpretable ML framework using Multi-Valued Decision Diagrams that works with missing data and includes invasive hemodynamics for risk stratifying advanced heart failure patients. In addition to producing a risk score, CARNA provides detailed patient phenotypes, i.e., sets of features and their thresholds that characterize a risk score.
\item We provide robust validation of the CARNA models using four independent HF cohorts, and compare them with six established HF risk scores and three traditional ML models. The CARNA models achieve high performance and outperform all benchmarks across metrics including Accuracy, Sensitivity, Specificity and AUC.
\item In order to facilitate practical use and promote open science, we provide an extensible, open-source tool implementation such that others can quickly and easily explore, extend, or prototype on top of the tool. In addition, our tool includes a deployed web server, which provisions live risk score prediction for ease of clinical use. All code is publicly available: \url{https://github.com/jozieLamp/CARNA}. 
% and the server may be accessed here: \url{http://hemopheno.pythonanywhere.com/}.
\end{enumerate}

% Intention that CARNA de
% Use of CARNA risk scores can facilitate clinical decision making about patient risk (e.g., which patient may be at higher risk for mortality or rehospitalization) enabling timely triage and treatment allocations.
% CARNA risk scores using explainable machine learning models can be used to identify advanced heart failure patient phenotypes which may be further investigated in future clinical research studies, e.g., elucidate the explicit effect and relationships between patient physiology and hemodynamics for heart failure symptoms.

\section{Related Work}

\subsection{Explainable AI in Healthcare}
Explainable AI (XAI) encompasses a wide range of techniques to provide interpretable explanations about a model's choices, such as through the use of feature maps, 
% (e.g., as used in as Shapley additive explanation (SHAP)), 
textual annotations, 
local, feature or example explanations, 
% local explanations via segmenting the solution space, 
and model simplification methods~\cite{arrieta2020explainable}. 
XAI methods have been used in a range of healthcare applications~\cite{bharati2023review}, including for predicting heart failure incidence~\cite{rao2022explainable}. 
% However, these methods typically provide explanations for black box ML models that cannot handle missing data. Furthermore, 
Adding explainability to a model often comes with a trade-off: the XAI methods may be complicated to implement, inefficient (i.e., it takes much longer to train an explainable model), or have stability and reliability issues~\cite{chaddad2023survey}. 
In addition, although these methods are good for understanding more about the model (e.g., visualizing at a high level how combinations of features contributed to a prediction), 
they are not always conducive to quick decision making in high stress environments. For example, interpreting feature maps or understanding textual explanations is nontrivial; one may need time to decipher the explanation and determine its use (e.g., figure out how a risk score was computed.)
On the other hand, MVDDs 
% support missing data, and 
are efficient, reliable and stable~\cite{florio2022optimal}. Moreover, they provide simple phenotypes that quickly summarize the exact set of features and thresholds that were used to make a score prediction. As such, they are easy to understand and provide an uncomplicated, quick decision support for clinicians during patient triage.

\begin{table}[t]
\caption{Comparison of HF Risk Score Approaches}\label{table:risk-score-comps}
\centering
\resizebox{\linewidth}{!}{%
\begin{tabular}{cccccc} \hline \toprule
\textbf{Score} & \textbf{Method Used} & \textbf{\# Features} & \textbf{Hemo?} & \textbf{Allows Missing Data}\\ \midrule
\textbf{CARNA Hemo} & MVDD & 28 & \textbf{Yes} & \textbf{Yes} \\
\textbf{CARNA All Fts}  & MVDD & 66 & \textbf{Yes} & \textbf{Yes} \\
EFFECT~\cite{lee2003predicting} & Logistic Regression & 11 & No & No \\
GWTG~\cite{peterson2010validated}  & Logistic Regression & 7 & No & No \\
MAGGIC~\cite{pocock2013predicting}  & Poisson Regression & 13 & No & No \\
ESCAPE~\cite{o2010escapescore}  & CPH & 8 & No & No \\
SHFM~\cite{levy2006seattle} & CPH & 30 & No & No \\
ADHERE~\cite{fonarow2005risk} & Decision Tree (CART) & 3 & No & No \\
MARKER~\cite{adler2020improving}  & Decision Tree (BDT) & 8 & No & No\\
TOPCAT~\cite{angraal2020machine}  & Various ML & 86 & No & No \\ \bottomrule
\end{tabular}
}
\caption*{\scriptsize Hemo = Invasive Hemodynamics; All Fts = All Features; CPH = Cox Proportional Hazards; CART = Classification and Regression Tree; BDT = Boosted Decision Tree}
\end{table}

\begin{table*}[t]
\caption{Characteristics of HF Cohorts}\label{table:patient-cohorts}
\centering
\resizebox{0.8\linewidth}{!}{%
\begin{tabular}{lccccc} \hline \toprule
 & \textbf{ESCAPE~\cite{binanay2005evaluation}} & \textbf{BEST~\cite{beta2001trial}} & \textbf{GUIDE-IT~\cite{felker2014rationale}} & \textbf{UVA Shock} & \textbf{UVA Serial} \\ \midrule
\# Patients & 433	& 2707	& 388	& 364	& 183 \\
\# Patients with Invasive Hemo & 209	& 0 &	0	& 130	& 181 \\
Baseline Data	& Yes	& Yes	& Yes	& Yes	& Yes \\
Discharge Data	& Yes	& No	& Yes	& Yes	& Yes \\
Total Records	& 866	& 2707	& 776	& 728	& 366 \\
Total Data Missing (\%) & 7.8	& 2.0	& 15.1	& 10.4	& 7.3 \\
Hemodynamics Missing (\%) & 12.0	& N/A	& N/A	& 5.9	& 9.2 \\
Age (years) & 56.1$\pm$13.9	& 60.2$\pm$12.3	& 62.2$\pm$13.9	& 59.4$\pm$18.5	& 60.6$\pm$15.1 \\
Sex (\% female) & 25.9	& 21.9	& 66.2	& 35.2	& 43.2 \\
Race (\% white) & 59.6	& 70.0	& 49.2	& N/A	& N/A \\
BMI (kg/m2)	& 28.4$\pm$6.7	& N/A	& 31.2$\pm$8.6	& 29.8$\pm$8.8	& 30.5$\pm$8.0 \\
LVEF (\%)	& 19.3$\pm$6.6	& 23.0$\pm$7.3	& 24.0$\pm$8.2	& 31.7$\pm$17.4	& 31.3$\pm$18.0 \\
SBP (mm Hg)	& 103.7$\pm$15.8	& 118.5$\pm$19.4	& 115.4$\pm$20.0	& 111.1$\pm$21.9	& 109.1$\pm$21.4 \\ 
DBP (mm Hg)	& 64.1$\pm$11.5	& 71.9$\pm$11.7	& 70.2$\pm$13.5	& 62.2$\pm$15.5	& 59.9$\pm$17.2 \\ 
Blood Urea Nitrogen (mg/dL)	& 36.3$\pm$22.5	& 24.6$\pm$15.3	& 31.3$\pm$22.6	& 34.9$\pm$24.2	& 39.1$\pm$25.7 \\
Creatinine (mg/dL)	& 1.5$\pm$0.6	& 1.2$\pm$0.4	& 1.6$\pm$0.7	& 1.7$\pm$1.3	& 1.7$\pm$1.0 \\
Potassium (mmol/L)	& 4.3$\pm$0.6	& 4.3$\pm$0.5	& 4.4$\pm$0.6	& N/A	& N/A \\
Sodium (mmol/L)	& 136.0$\pm$4.4	& 138.9$\pm$3.4	& 138.3$\pm$3.8	& 136.9$\pm$5.1	& 135.7$\pm$5.2 \\
DeLvTx (\%) & 27.0	& 31.7	& 23.7 & 56.6	& 41.5 \\
Rehospitalization (\%) & 57.0	& 62.9	& 51.8	& 47.5 & 78.7 \\ \bottomrule
\end{tabular}
}
\parbox{0.8\linewidth}{\caption*{\scriptsize N/A indicates data not available; LVEF = ejection fraction; SBP = systolic blood pressure; DBP = diastolic blood pressure; DeLvTx = composite endpoint of death, LVAD  implantation or transplantation.}}
\end{table*}

\subsection{HF Risk Scores}
% 6 other established HF risk scores: ADHERE~\cite{fonarow2005risk}, EFFECT~\cite{lee2003predicting}, ESCAPE~\cite{o2010escapescore}, GWTG~\cite{peterson2010validated}, MAGGIC~\cite{pocock2013predicting}, and SHFM~\cite{levy2006seattle}.
There are a variety of HF risk scores that provide risk stratifications  
% predict risk of mortality 
in HF populations using statistical and machine learning models; a comparison is available in Table~\ref{table:risk-score-comps}. 
% decision trees, regression methods and cox proportional hazards models. 
% The Enhanced Feedback for Effective Cardiac Treatment (EFFECT) 
The EFFECT~\cite{lee2003predicting}, 
% American Heart Association's Get With the Guidelines–Heart Failure 
GWTG~\cite{peterson2010validated} and  
% Meta-analysis Global Group in Chronic Heart Failure (MAGGIC) 
MAGGIC~\cite{pocock2013predicting} risk scores predict risk of mortality in HF patients using various regression methods.
The ESCAPE 
% (Evaluation Study of Congestive Heart Failure and Pulmonary Artery Catheterization Effectiveness) 
Risk Model and Discharge Score~\cite{o2010escapescore} and 
% the Seattle Heart Failure Model 
SHFM~\cite{levy2006seattle} stratify mortality risk using Cox Proportional Hazards models (CPH). The ESCAPE score was derived using the same dataset that we use for our training cohort. 
Finally, in the TOPCAT~\cite{angraal2020machine},
% in Angraal et al.~\cite{angraal2020machine} using the Treatment of Preserved Cardiac Function Heart Failure with an Aldosterone Antagonist (TOPCAT) trial, 
% the Acute Decompensated Heart Failure National Registry (ADHERE) Risk Score
ADHERE~\cite{fonarow2005risk}, 
% and the Machine learning Assessment of RisK and EaRly mortality in Heart Failure (MARKER) risk model
and MARKER~\cite{adler2020improving} risk models, machine learning algorithms, including decision trees, boosted decision trees, support vector machines and random forests are used to predict risk of mortality. 

Some of these risk models use small, selective feature sets, or only stratify risk into a small number of groups (e.g., only two groups of high and low risk as in MARKER), and none of them incorporate invasive hemodynamics. 
%incorporate invasive hemo
Moreover, these methods suffer from limitations associated with statistical and naive machine learning models, such as being prone to bias, and lacking mechanisms to handle missing data~\cite{di2020evaluating,canepa2018performance}. 
% Moreover, none of them are able to handle missing data; 
% in external validation studies using these scores, a common issue cited is that some
In fact, in external validation of these scores, a common issue cited is that some variables 
% In fact, a common issue cited in external validation studies using these scores is that some variables 
are not readily available in routine clinical practice or are missing from collected data cohorts so the score cannot be computed~\cite{codina2021head}.

CARNA uses a larger, more diverse feature set than most scores, is able to provide more fine-grained risk stratification, i.e., can have more risk groups, and incorporates invasive hemodynamics. In addition, our model is explainable and can handle missing data. 
% Our score also outperforms others in our risk predictivity 
Ultimately, it is our intention that our risk score would be complementary to previous risk methodologies, in which our score is used to provision risk stratification for advanced HF patients requiring invasive hemodynamic monitoring, and others may be used to gain an understanding of risk for more general HF patients.

\section{Preliminaries}
\subsection{Multi-Valued Decision Diagrams}
MVDDs are discrete structures representing logical functions in directed, acyclic graphs where nodes represent features, edges represent logical operators (“and”, “or”) with parameter threshold values, and leaf nodes represent the final score classification~\cite{srinivasan1990algorithms}. As such, the “path” through the graph may be returned to provide a descriptive patient phenotype. An example MVDD is shown in Figure~\ref{fig:example-mvdd}: the highlighted red path characterizes the high-risk score of 5 by the following phenotype: 
\textit{Sex = Male $\land$ BPSYS $>$ 103.5 $\land$ CPI $>$ 0.621 $\land$ (PAS $>$ 74.5 $\lor$ PCWP $\leq$ 33) = Score 5}.

MVDDs are well suited to classification tasks and the representation of HF phenotypes over other black-box models because they allow increased flexibility in characterizing feature relationships and are highly interpretable~\cite{florio2022optimal}. This is advantageous over other models that do not provide any details about how a score was computed. Moreover, unlike other explainable models such as decision trees or random forests, MVDDs are resilient to missing data due to their use of logical operators; multiple substitutable features may contribute to the same prediction score. For example, in the above phenotype, PAS or PCWP may be used for calculation, and as such, when a feature is missing from the provided data, alternative features may be used to still allow for score prediction. This is advantageous in clinical scenarios where complete patient measurements may not be available and clinicians must make quick decisions on partial observations.

Despite these advantages, MVDDs have typically been used for optimization and model checking contexts~\cite{bergman2016decision}, with limited use in medical classification and no applications to risk stratification. As such, we develop a training regime for MVDDs within our framework to learn risk scores and output HF phenotypes that characterize the predicted risk scores.

\begin{figure*}[t]
    \centering
    \includegraphics[width=\linewidth]{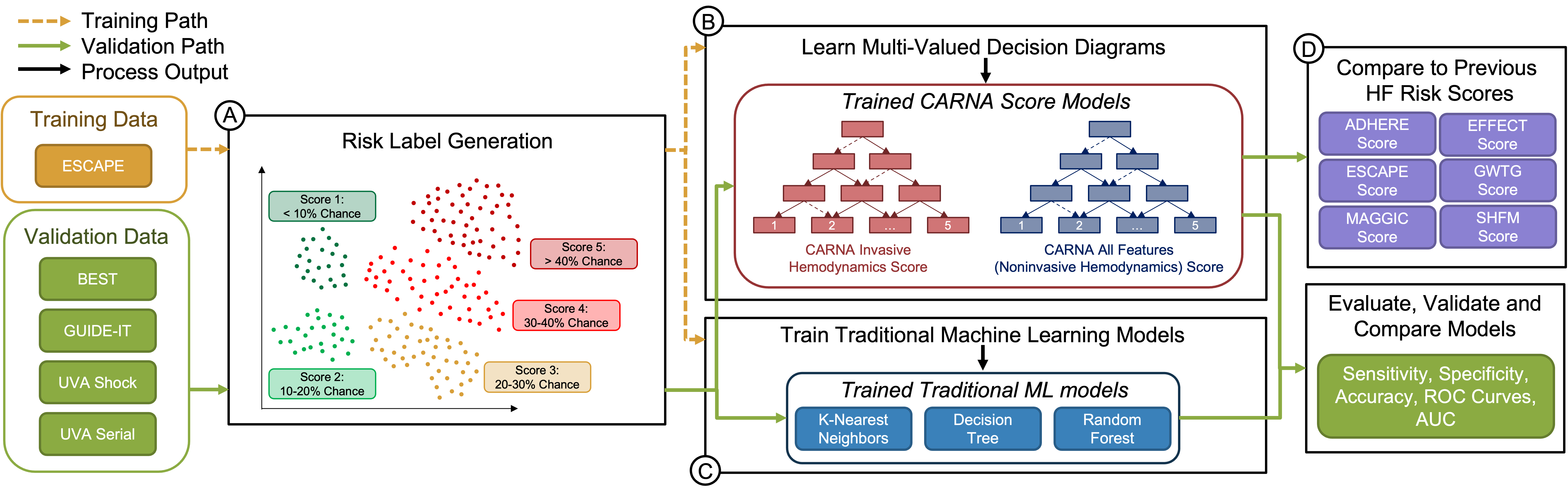}
    \caption{Overview of CARNA Methodology. (A) The risk labels are generated using a clustering-based derivation scheme using the training and validation datasets; (B) the training data is used to train the CARNA MVDD models as well as (C) three traditional ML models for comparison. Finally, the validation data is used to evaluate the performance of the models and the resulting CARNA risk scores are compared with six previous HF risk scores (D). }
    \label{fig:overview}
\end{figure*}

\section{Data}
\subsection{Outcomes and Cohort Selection}
\subsubsection{Outcomes}
The primary outcome was a composite endpoint of death, left ventricular assist device (LVAD) implantation or heart transplantation (denoted as DeLvTx). A secondary outcome of rehospitalization within 6 months of follow up was included, as rehospitalizations have been shown to be predictive of adverse outcomes~\cite{ahmed2006higher,desai2012rehospitalization}. 

\subsubsection{Patient Cohorts}
This study used 5 HF cohorts, three from randomized clinical trials and two from a real-world setting of a single quaternary healthcare system. We trained the model using the ESCAPE 
% (Evaluation Study of Congestive Heart Failure and Pulmonary Artery Catheterization Effectiveness) 
trial [433 patients, mean age 56.1, 25.9\% female], a randomized control trial studying the use of pulmonary artery catheters in severe HF patients~\cite{binanay2005evaluation}. The ESCAPE dataset contains a rich feature set of clinical and hemodynamic variables. Invasive hemodynamics (e.g., right atrial pressure (RAP) and pulmonary capillary wedge pressure (PCWP)) were recorded for 209 patients at baseline and prior to the removal of a heart catheter. The other 4 cohorts were used for validation: 
% The Beta-Blocker Evaluation of Survival Trial (BEST) [2707 patients, mean age 60.2, 21.9\% female], was a randomized control trial that tested whether bucindolol hydrochloride reduced mortality among HF patients~\cite{beta2001trial}. The Guiding Evidence Based Therapy Using Biomarker Intensified Treatment in Heart Failure (GUIDE-IT) [894 patients, mean age 61.5, 68\% female] trial was a randomized controlled unblinded trial testing the efficacy and safety of adjusting therapy to maintain a N-terminal pro–B-type natriuretic peptide level of less than 1000 pg/ml in systolic HF patients~\cite{felker2014rationale}. We also performed external validation on two additional real-world cohorts from the University of Virginia: 1) a registry of cardiogenic shock patients [364 patients, mean age 59.4, 11.7\% female], and 2) a registry of HF patients who had at least two serial right heart catheterizations for hemodynamic assessment during the same hospitalization [183 patients, mean age 60.6, 43.2\% female]. 
% Beta-Blocker Evaluation of Survival Trial (BEST)~\cite{beta2001trial}, Guiding Evidence Based Therapy Using Biomarker Intensified Treatment in Heart Failure (GUIDE-IT) trial~\cite{felker2014rationale} 
BEST~\cite{beta2001trial}, GUIDE-IT~\cite{felker2014rationale} 
and two real-world cohorts from University of Virginia (UVA); 1) a registry of cardiogenic shock patients, and (UVA Shock) and 2) a registry of HF patients with at least two serial right heart catheterizations for hemodynamic assessment during the same hospitalization (UVA Serial). More details are available in the cohort publications~\cite{binanay2005evaluation, beta2001trial, felker2014rationale}. 
Only New York Heart Association (NYHA) functional class III-IV were included in the study to ensure comparability. Characteristics of each cohort are in Table~\ref{table:patient-cohorts}. 
Only the ESCAPE, UVA Cardiogenic Shock and UVA Serial Cardiac cohorts have invasive hemodynamics. GUIDE-IT had the highest percentage of missing data (15.07\%), and ESCAPE had the highest percentage of missing hemodynamic data (12.04\%). 
% Additional characteristics for each of the cohorts, including the percentage of missing data, are included in Table~\ref{table:patient-cohorts}. 
% This study has been approved by the University of Virginia Institutional Review Board. The ESCAPE, BEST and GUIDE-IT data are available to other researchers for purposes of reproducing the results or replicating the procedure via data request from the National Heart, Lung, and Blood Institute Biological Specimen and Data Repository Information Coordinating Center. The UVA cohorts are available from the corresponding author upon reasonable request.
% Additional characteristics for each of the cohorts, including the percentage of missing data and outcome occurences are included in Table~\ref{table:patient-cohorts}. 
% Outcome characteristics for each cohort are shown in Table~\ref{table:patient-cohorts}.

\subsection{Data Preprocessing}
For each dataset, we first preprocessed the data, including removing outliers (necessary to reduce bias in our ML training). If the dataset had multiple temporal recordings, the values recorded at baseline and discharge were treated as two separate records. Baseline values were included (as opposed to only discharge) as they have been shown to inform a range of hemodynamic and contractile metrics and are also important in predicting outcomes as shown in previous studies, e.g.,~\cite{bilchick2018clinical}. Moreover, since it was the authors’ intention to provide a single point of care risk score that could make predictions even at initial hospital admission, we included baseline measurements in the models. This also increased the total number of training/validation records, especially helpful with the small training (ESCAPE) dataset. Which cohorts had baseline and discharge data, as well as the total number of records used is reported in Table~\ref{table:patient-cohorts}.

Since our models support missing features, we did not impute or remove missing values from the data records. We also calculated noninvasive hemodynamics, additional metrics indicative of hemodynamic states, computed from features that were collected noninvasively. Examples include mean arterial pressure (MAP), 
% mean pulmonary artery pressure (MPAP), 
cardiac power index (CPI), and pulse pressure (PP). These metrics were specifically selected a priori based on previous studies demonstrating incremental value in HF risk stratification~\cite{bilchick2018clinical,bilchick2018plasma,mazimba2016decreased}. Data was stratified into two subsets: one exploring phenotypes of invasive hemodynamics only, and the other for characterizing phenotypes between noninvasive hemodynamics and all available clinical variables including demographics, labs, and medications. Henceforth, we refer to these as the Invasive Hemodynamics and All Features feature sets, respectively.
\section{Methods}

% TODO HERE- update to have sectino refs of a b c d and update overview figure also ... 
% \subsection{Method Overview}
A high-level overview of the CARNA methodology is shown in Figure~\ref{fig:overview}. First, the risk labels are generated (Section~\ref{sec:methods-risk-label}). Agglomerative clustering is used to stratify patients in all datasets into a specified number of cluster groups and risk categories are derived for each cluster (e.g., class 1-5 ordered numerically based on actual event rates). The output of this step is a set of risk labels that indicate the probability threshold of the outcome event happening (e.g., a patient record assigned to a class of 1 indicates an outcome probability of $<$10\% for that patient). This clustering occurred twice (once for each feature set), and the probabilities from each cluster were derived for each outcome, resulting in a total of four risk label sets. Next, using the training data (ESCAPE cohort), Multi-Valued Decision Diagrams were trained to predict the risk labels (e.g., classes 1-5, Section~\ref{sec:methods-learn-mvdd}). 
The trained MVDD models take in a set of features for a patient and output the predicted CARNA risk score. A total of four models were derived for each of the four risk label sets: one for each outcome (DeLvTx, Rehospitalization) and feature set (Invasive Hemodynamics, All Features) pair. Finally, the CARNA risk scores were evaluated using the four other validation cohorts and compared with traditional ML models (Section~\ref{sec:methods-trad-ml}) and established HF risk scores (Section~\ref{sec:methods-hf-risk-score}) based on their predictions of the risk classes. A step-by-step walkthrough of the methodology is provided next.

\subsection{Risk Label Generation}\label{sec:methods-risk-label}
Each patient cohort had binary outcomes for the two endpoints (DeLvTx, rehospitalization), indicating if the outcome occurred or not. As such, there were no explicit risk thresholds for each of the patient records. Additionally, our MVDDs do not implicitly assign risk scores as a function of their learning. Since the goal of our approach was to generate a risk stratification and phenotyping score, the next step was to generate the categorical risk score values (i.e., 1–5) corresponding to real-valued outcome risks (e.g., 1 indicates a $<$10\% risk of DeLvTx) for each record in the training and validation datasets. To this end, we reduced the dimensions of the covariates in the datasets using Principal Component Analysis (PCA) and then used a clustering approach to group patients and determine risk categorizations. 

\subsubsection{PCA} For each feature set and outcome, we performed PCA using two principal components to reduce the dimensions of the data. This was a necessary pre-step to reduce bias in the clustering, as clustering methods can be sensitive to outliers or slight changes in feature set distributions. Specifically, we use the LAPACK implementation of Singular Value Decomposition, following the PCA library available in the scikit-learn packages~\cite{pedregosa2011scikit}. 
% In this library, features with the highest variances are selected as the principal components. 
Since PCA cannot handle missing features, we imputed any missing values with the feature mean. We note however, that the original (non-imputed) datasets were used in the MVDD training steps later to ensure the models learned from the datasets with missing data. Importantly, the risk scores generated from this step were the labels used to train the MVDD models.

\subsubsection{Hierarchical Clustering} Next, we clustered the patients into a specified number of groups using Agglomerative Clustering, a form of hierarchical clustering. 
Since the number of groups, $k$, is a hyperparameter, the users can select how many groups they wish to stratify the patients into. We argue this is an advantage of our approach because, based on details of the patient cohort being trained on or other user criteria (e.g., a clinician wish for only three risk groups), the number of risk groups can be adaptively selected. For our experimental purposes, we selected $k$ as the optimal number of groups using the 
% To choose the optimal number of groups, we used the 
``Elbow” method, in which the sum of squares at each number of clusters is plotted on a graph~\cite{tibshirani2001estimating}. The point on the graph where the slope changes from steep to shallow (``elbow” of the graph) indicates the optimal number of clusters to use. The clustering was performed across all datasets (including the four validation cohorts). In order to discriminate how well separated the clusters were, we computed the Hubert \& Levin C Index for each feature set~\cite{hubert1976general}. The C Index provides a metric to compare the dispersion of clusters compared to the overall dataset dispersion~\cite{pypi}. C Index should be minimized; a smaller index indicates more distinct (stable) clusters. Each cluster corresponds to one score value (e.g., five clusters for five score value assignments).

\subsubsection{Derive Outcome Probabilities} From there, the outcome probability ranges for each score cluster were derived by computing the ground truth probability of the denoted outcome from the patients in each cluster. For example, cluster 1, corresponding to a score value of 1, had a ground truth probability of 0.041 for the DeLvTx outcome and Invasive Hemodynamics feature set; an outcome probability of $<$10\% was derived. As a sanity check to ensure the derived score categories corresponded to the ground truth outcome probabilities across all the datasets, we reported the actual probabilities for each dataset in the Results. Finally, the score labels were assigned to each data record based on the associated cluster (e.g., a record in cluster 1 is assigned a score of 1). Using this process, we generated the risk score (labels) separately for each outcome and feature set, resulting in a total of four risk score label sets. 

\subsubsection{Label Method Reasoning} We decided to use this clustering approach because, in addition to risk stratifying patients, 
it uses an unsupervised method to holistically group patients, 
% it holistically groups patients in an unsupervised manner, 
i.e., autonomously groups patients based on similar characteristics. This is highly advantageous over manually stratifying patients; manually grouping patients into risk groups is nontrivial due to large (potentially conflicting) sets of features and high variability in the presentation of patient conditions. Moreover, manual grouping is labor intensive (e.g., would require many clinician hours to characterize every patient’s risk).

\subsection{Learning Multi-Valued Decision Diagrams}\label{sec:methods-learn-mvdd}

\subsubsection{Overall Training Details}
As a reminder, the MVDDs were trained on the risk score labels (i.e., classes 1–5) generated during the previous step and the risk score labels indicate probability categories of outcomes.
The resulting trained models take in a set of features for a patient and output the predicted CARNA risk score. 
All MVDDs were learned using an independent training set (ESCAPE dataset). 
To maximize the training capabilities of the small dataset, we used 5-fold cross validation, in which 80\% of the data in the split was used for training and the other 20\% was held out for validation purposes. A total of four models were derived: one for each outcome (DeLvTx, Rehospitalization) and feature set (Invasive Hemodynamics and All Features) pair.

\subsubsection{MVDD Learning Process}
Each MVDD was learned using a training process similar to the Iterative Dichotomiser 3 (ID3) multi-class decision tree algorithm~\cite{quinlan1986induction}. Specifically, we learn a multi-class tree using the splitting criterion of gini index or entropy. Each time we add a node to the tree, we replace the boolean edge with logical operators (``and", ``or") and select the operator that gives the best performance (e.g., lowest gini or entropy.) The MVDDs were trained iteratively until model convergence. 
% After each iteration, the predicted set of classes was compared with the actual set of classes, enabling the MVDD to learn and improve with each iteration. 
The implementation was developed de novo in Python3 using publicly available packages~\cite{pedregosa2011scikit}.

\subsubsection{Validating the MVDDs} After model training, we independently validated the models using the four other cohorts, which had not been used in the training phase. To assess the performance of our MVDDs, five receiver operator characteristic curves (ROC) for each risk class were plotted for each model based on the ground truth risk classes in the validation datasets. If the predicted risk class matched the ground truth risk class, this was considered a success for the ROC analysis. For example, in the case of class 1 patients, if the MVDD predicted class 1, it was considered a success, and if it predicted another class, it was considered a failure. The ROC curves were then constructed based on predictions of the risk classes, which is different from the conventional ROC method of predicting an actual event.
To measure the overall model performance (e.g., as a summary metric across all risk classes,) we report a single averaged area under the curve (AUC) metric, calculated by taking the weighted average of the AUCs from each risk class, weighted by the number of individuals in each class. We also calculated accuracy, sensitivity and specificity in a similar manner. We note that ROC/AUC were used over a reclassification analysis due to limitations associated with reclassification such as systematic miscalibration on validation cohorts~\cite{leening2014net}. 

\begin{table*}[t]
\caption{Risk Score Meaning and Ground Truth Risk Probabilities}\label{table:risk-cats}
\centering
\caption*{DeLvTx Outcome}
\resizebox{\linewidth}{!}{%
\begin{tabular}{ccc|cccc|cccccc} \toprule
& & & \multicolumn{4}{c}{\textbf{Invasive Hemodynamics Cluster Means}} & \multicolumn{6}{|c}{\textbf{All Features Cluster Means}} \\
\textbf{Risk Score} & \textbf{Probability} & \textbf{Risk Category} & \textbf{Overall} & \textbf{ESCAPE} & \textbf{UVA Shock} & \textbf{UVA Serial} & \textbf{Overall} & \textbf{ESCAPE} & \textbf{BEST} & \textbf{GUIDE-IT} & \textbf{UVA Shock} & \textbf{UVA Serial}\\ \midrule
1 & \textless 10\% & Low & 0.041 & 0.081 & N/A & 0.0 & 0.043 & 0.042 & 0.0 & 0.076 & 0.048 & 0.048 \\
2 & 10 - 20\% & Low - Intermediate & 0.176 & 0.185 & N/A & 0.167 & 0.145 & 0.129 & 0.159 & 0.143 & 0.167 & 0.125 \\
3 & 20 - 30\% & Intermediate & 0.245 & 0.25 & 0.227 & 0.259 & 0.255 & 0.265 & 0.275 & 0.235 & 0.201 & 0.299 \\
4 & 30 - 40\% & Intermediate - High & 0.364 & 0.39 & 0.31 & 0.392 & 0.343 & 0.333 & 0.331 & 0.253 & 0.315 & 0.485 \\
5 & \textgreater 40\% & High & 0.535 & 0.429 & 0.651 & 0.525 & 0.688 & 0.769 & 0.333 & 0.338 & 1.0 & 1.0 \\ \bottomrule
\end{tabular}
}
\hfill
\caption*{Rehospitalization Outcome}
\resizebox{\linewidth}{!}{%
\begin{tabular}{ccc|cccc|cccccc} \toprule
& & & \multicolumn{4}{c}{\textbf{Invasive Hemodynamics Cluster Means}} & \multicolumn{6}{|c}{\textbf{All Features Cluster Means}} \\
\textbf{Risk Score} & \textbf{Probability} & \textbf{Risk Category} & \textbf{Overall} & \textbf{ESCAPE} & \textbf{UVA Shock} & \textbf{UVA Serial} & \textbf{Overall} & \textbf{ESCAPE} & \textbf{BEST} & \textbf{GUIDE-IT} & \textbf{UVA Shock}& \textbf{UVA Serial}\\ \midrule
1 & \textless 10\% & Low & 0.025 & 0.05 & N/A & 0.0 & 0.035 & 0.077 & 0.05 & 0.017 & 0.0 & 0.031 \\
2 & 10 - 20\% & Low - Intermediate & 0.102 & 0.203 & N/A & 0.0 & 0.163 & 0.186 & 0.125 & 0.177 & 0.173 & 0.156 \\
3 & 20 - 30\% & Intermediate & 0.261 & 0.276 & 0.216 & 0.291 & 0.286 & 0.309 & 0.275 & 0.259 & 0.275 & 0.312 \\
4 & 30 - 40\% & Intermediate - High & 0.379 & 0.407 & 0.431 & 0.30 & 0.342 & 0.333 & 0.312 & 0.405 & 0.332 & 0.328 \\
5 & \textgreater 40\% & High & 0.779 & 0.647 & 0.798 & 0.892 & 0.724 & 0.667 & 0.632 & 0.571 & 0.75 & 1.0 \\ \bottomrule
\end{tabular}
}
\caption*{\scriptsize Tables display risk scores with corresponding outcome probability ranges and risk categories as well as cluster means, the ground truth mean outcome probability for each risk cluster in each dataset. Overall is the ground truth mean outcome probability across the entire cluster (i.e., across all datasets in the cluster). N/A = no data points assigned to that cluster; DeLvTx = composite endpoint of death, LVAD implantation or transplantation. 
% CM = Cluster Mean; CM is the ground truth mean probability of the outcome for each risk cluster. 
% Overall Cluster Mean is the ground truth mean probability of the outcome across the entire cluster (i.e., across all datasets in the cluster).
}
\end{table*}

% \begin{figure*}
%      \centering
%       \subfigure[Invasive Hemodynamic]{ 
%          \centering
%          \includegraphics[width=0.43\linewidth]{Figures/Dendogram_Hemo.png}}
%         \hfill
%      \subfigure[All Features]{
%          \centering
%          \includegraphics[width=0.43\linewidth]{Figures/Dendogram_AllData.png}}
%     \caption{Agglomerative Clustering Dendograms for the Invasive Hemodynamic (a) and All Features (b) feature sets. 
%     % Resulting dendrograms displaying the cluster splits are shown for the Invasive Hemodynamic (a) and All Features (b) feature sets. 
%     Clusters are separated by horizontally dividing the top of the hierarchy based on the number of groups $k$ (5 in our case), illustrated by the horizontal dashed line in the figures. Each leaf (end of the dendrogram) represents an individual data point.}
%     \label{fig:dendograms}
% \end{figure*}

\begin{figure}[t]
    \centering
    \includegraphics[width=0.95\linewidth]{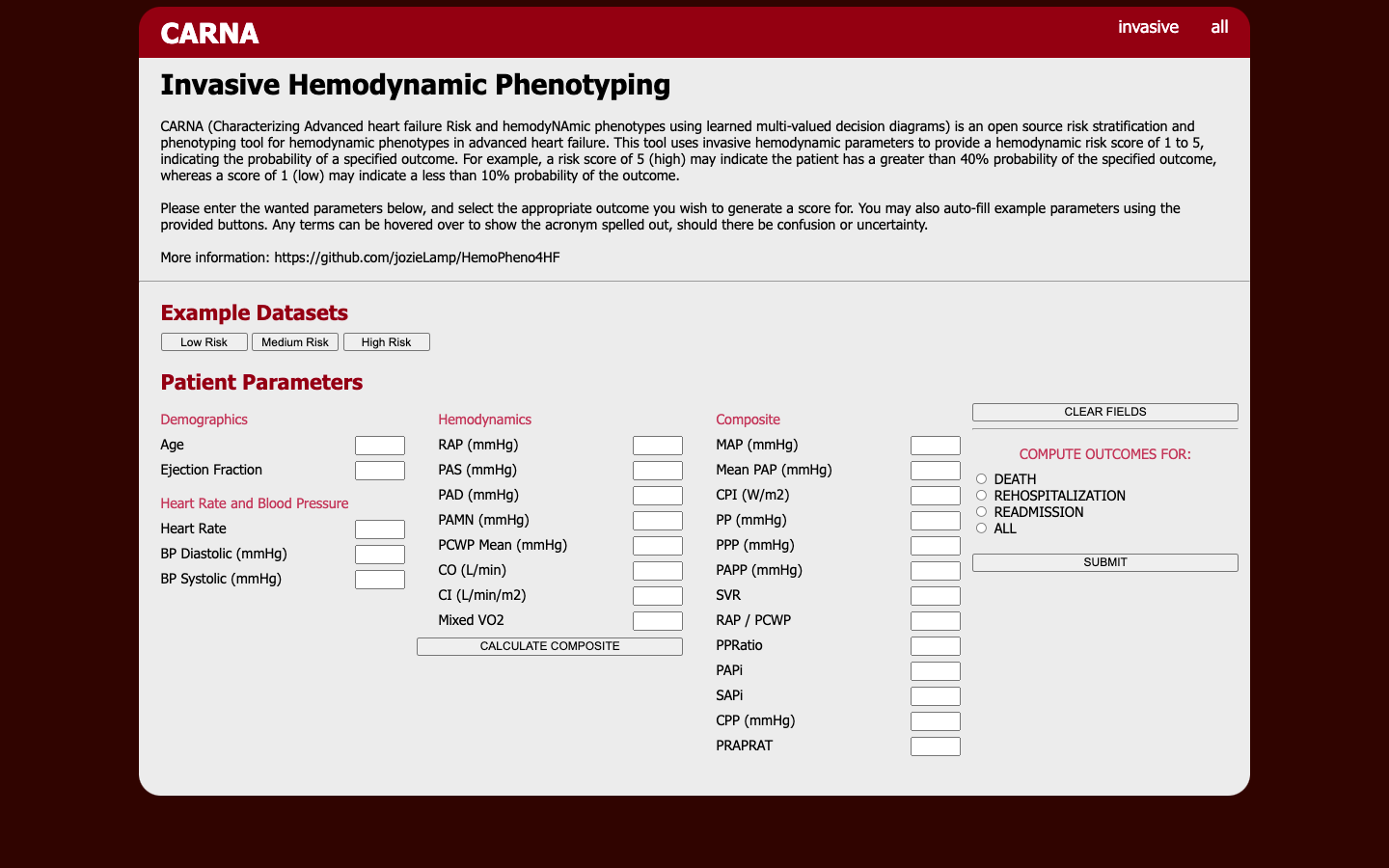}
    \caption{Example CARNA Web Portal -- interface for predicting the invasive hemodynamic risk score. }
    \label{fig:web-portal}
\end{figure}

\subsection{Comparison to Traditional Machine Learning Models}\label{sec:methods-trad-ml}
We compared the performance of CARNA models with traditional ML models, including K-nearest neighbors (KNN), Decision Trees (DT) and Random Forests (RF). Median imputation was used for any missing values. We followed the same training procedure used for the MVDDs; each model was trained on the ESCAPE dataset using 5-fold cross validation with a 80-20\% split for training/validation. 
% Validation was done using the four other cohorts and performance was reported using the same metrics. 
Performance was computed using the same metrics on the four validation cohorts. 
Additionally, to assess the concordance between the predicted risk 
% categories 
and the ground truth outcomes, calibration plots were computed, using a bin size of 10.

\subsection{Comparison to Other Heart Failure Risk Scores}\label{sec:methods-hf-risk-score}
For benchmark comparison, we compared our CARNA risk score models with six other established HF risk scores: ADHERE~\cite{fonarow2005risk}, EFFECT~\cite{lee2003predicting}, ESCAPE~\cite{o2010escapescore}, GWTG~\cite{peterson2010validated}, MAGGIC~\cite{pocock2013predicting}, and SHFM~\cite{levy2006seattle}. 
We limited our comparison to the models predicting risk of mortality with similar feature sets and patient cohorts. In particular, we exclude scores that use biomarkers and pathology based features (e.g., QRS measurements) since those were not available in our cohorts.
Since the comparison scores cannot handle missing data, missing values were imputed with the median. For each validation dataset, the predicted probability of an event was obtained from each score for each patient, and then a predicted class was assigned based on that probability. In other words, if the predicted probability of the event from the SHFM was 5\% for a patient, we would say the SHFM predicted class 1, which had a probability range of 0-10\% for an event. The accuracy of these other models for predicting the risk class (not the actual event) was again used for the comparison ROC analysis. 
% ~\cite{pedregosa2011scikit}. 
To compare the AUCs between the established HF risk scores and CARNA, we performed hypothesis testing using the DeLong approach~\cite{delong1988comparing}. We report the scores’ AUCs, the change in AUCs (CARNA AUC – other score AUC) and the p-value.

\subsection{Open Source Tool Implementation}\label{sec:methods-tool}
In order to promote open science, CARNA is an open source, extensible framework that others can easily use and build off of. Our implementation is developed in Python 3 using open source libraries. The tool package is clearly commented and includes a jupyter notebook runner file such that others can quickly and easily explore, extend, or prototype on top of the tool. In addition, our implementation includes a deployed web server which provides a live risk score prediction for ease of clinical use. An example web portal image is in Figure~\ref{fig:web-portal}. All code is publicly available from the Github repository: \url{https://github.com/jozieLamp/CARNA}, and the live web server may be accessed here: \url{http://hemopheno.pythonanywhere.com/}.

\begin{figure}
    \centering
    \includegraphics[width=\linewidth]{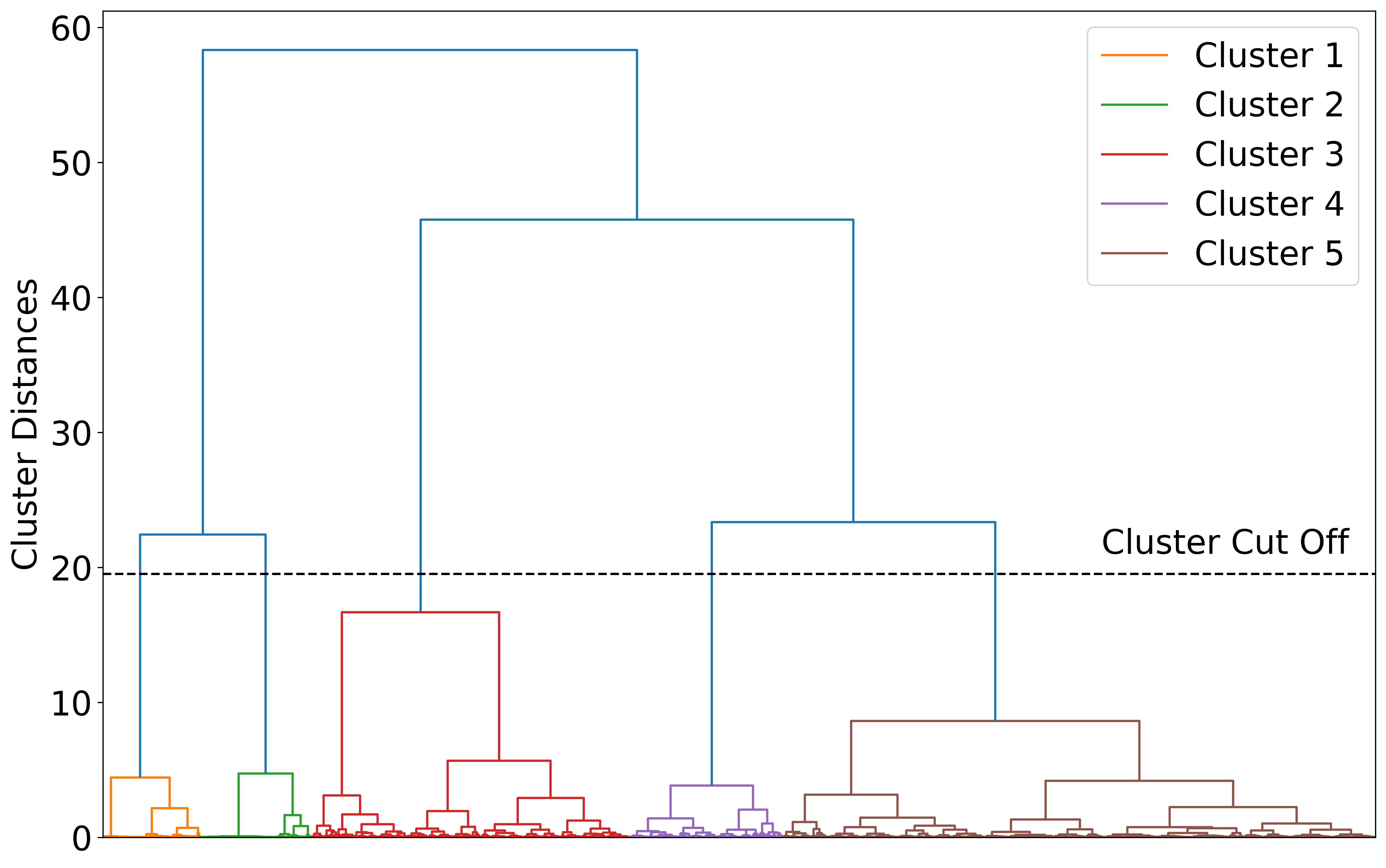}
    \caption{Agglomerative Clustering Dendogram for All Features feature set. Clusters are separated by horizontally dividing the top of the hierarchy based on the specified number of groups (5 in our case); this is illustrated by the horizontal dashed black line in the figures. Each leaf (end of the dendrogram) represents an individual data point.}
    \label{fig:dendograms}
\end{figure}

\section{Results} %Experimental Results

% \subsection{Patient Characteristics}
% Baseline characteristics for each cohort are available in Table~\ref{table:patient-cohorts}. Only the ESCAPE, UVA Cardiogenic Shock and UVA Serial Cardiac cohorts had invasive hemodynamics. GUIDE-IT had the highest percentage of missing data (15.07\%), and ESCAPE had the highest percentage of missing hemodynamic data (12.04\%). 
% % More details are available in the cohort publications~\cite{binanay2005evaluation, beta2001trial, felker2014rationale}.

\begin{table*}[t]
\caption{Model Performance Summary (Validation Data)}\label{table:carna-performance}
\centering
\caption*{Invasive Hemodynamic Feature Set}
\resizebox{0.8\linewidth}{!}{%
\begin{tabular}{cccccc} \hline \toprule
\textbf{Outcome} & \textbf{Dataset} & \textbf{Accuracy} & \textbf{Averaged AUC} & \textbf{Sensitivity} & \textbf{Specificity} \\ \midrule
\multirow{2}{*}{DeLvTx} & UVA Shock & 0.947$\pm$0.107	& 0.938$\pm$0.106	& 0.915$\pm$0.103	& 0.961$\pm$0.108 \\ 
& UVA Serial & 0.969$\pm$0.081	& 0.965$\pm$0.080 &	0.950$\pm$0.079	& 0.980$\pm$0.081 \\ \midrule
\multirow{2}{*}{Rehospitalization} & UVA Shock & 0.907$\pm$0.102 & 0.861$\pm$0.096	& 0.791$\pm$0.086	& 0.935$\pm$0.105 \\
& UVA Serial & 0.896$\pm$0.074 &	0.896$\pm$0.074	& 0.852$\pm$0.070	& 0.940$\pm$0.078 \\ \bottomrule
\end{tabular}
}
\vspace*{\baselineskip}
\caption*{All Features Feature Set}
\resizebox{0.8\linewidth}{!}{%
\begin{tabular}{cccccc} \hline \toprule
\textbf{Outcome} & \textbf{Dataset} & \textbf{Accuracy} & \textbf{Averaged AUC} & \textbf{Sensitivity} & \textbf{Specificity} \\ \midrule
\multirow{4}{*}{DeLvTx} & BEST & 0.997$\pm$0.037	& 0.994$\pm$0.037	& 0.990$\pm$0.037	& 0.998$\pm$0.037 \\
& GUIDE-IT & 0.997$\pm$0.070	& 0.996$\pm$0.070	& 0.995$\pm$0.069	& 0.998$\pm$0.070 \\
& UVA Shock & 0.865$\pm$0.049	& 0.871$\pm$0.050	& 0.811$\pm$0.045	& 0.931$\pm$0.054 \\
& UVA Serial & 0.858$\pm$0.067	& 0.871$\pm$0.068	& 0.815$\pm$0.063	& 0.927$\pm$0.073 \\ \midrule
\multirow{4}{*}{Rehospitalization} & BEST & 0.997$\pm$0.037	& 0.994$\pm$0.037	& 0.990$\pm$0.037	& 0.998$\pm$0.037 \\
& GUIDE-IT & 0.997$\pm$0.070	& 0.996$\pm$0.070	& 0.995$\pm$0.069	& 0.998$\pm$0.070 \\
& UVA Shock & 0.694$\pm$0.036 & 0.533$\pm$0.015	& 0.250$\pm$0.041 & 0.816$\pm$0.046 \\
& UVA Serial & 0.890$\pm$0.070	& 0.798$\pm$0.061	& 0.653$\pm$0.044	& 0.942$\pm$0.075 \\ \bottomrule
\end{tabular}
}
\parbox{0.8\linewidth}{\caption*{\scriptsize Table displays value $\pm$ confidence interval. DeLvTx = composite endpoint of death, LVAD implantation or transplantation.}}
\end{table*}

\begin{figure*}[t]
    \centering
    \includegraphics[width=\linewidth]{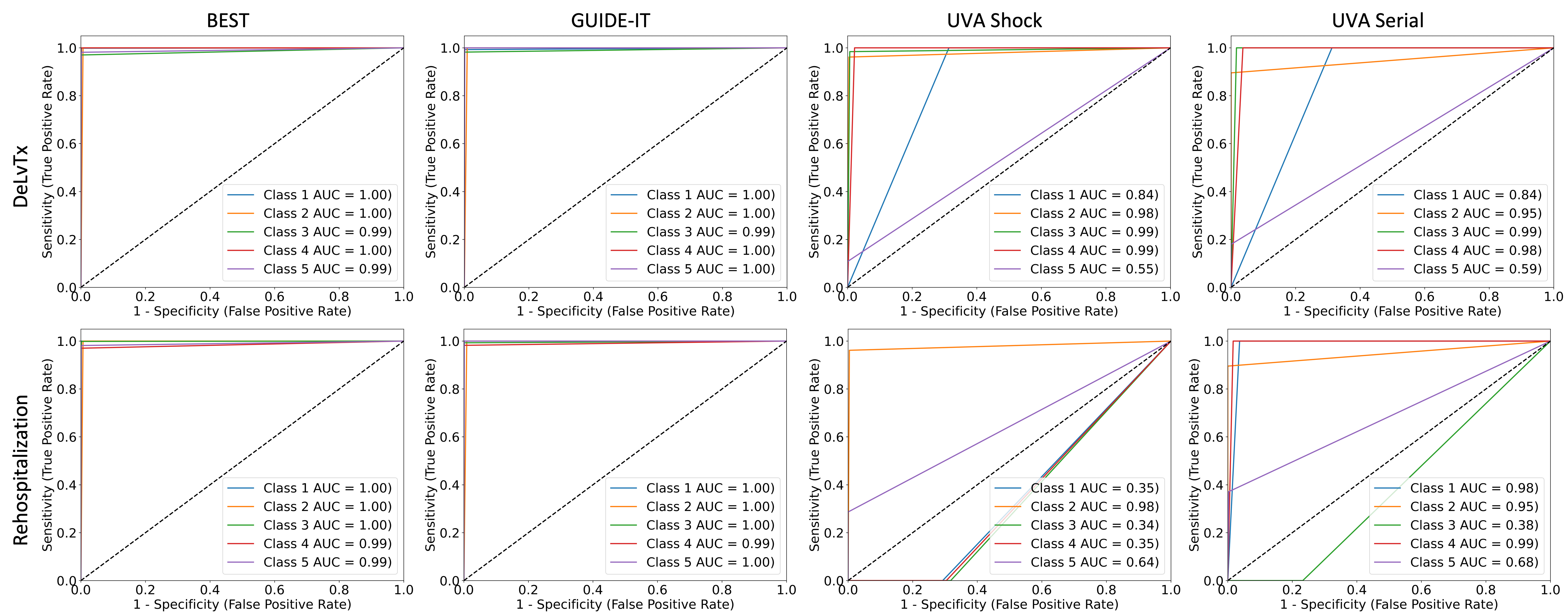}
    % \vspace{-0.6cm}
    \caption{ROC Curves for Validation Datasets and All Features Feature Set.}
    % . nan indicates no data in that class; DeLvTx, composite endpoint of death, left ventricular assist device (LVAD) implantation or transplantation.}
    % \vspace{-0.6cm}
    \label{fig:carna-allfts-roc}
\end{figure*}

\begin{figure*}[t]
    \centering
    \includegraphics[width=\linewidth]{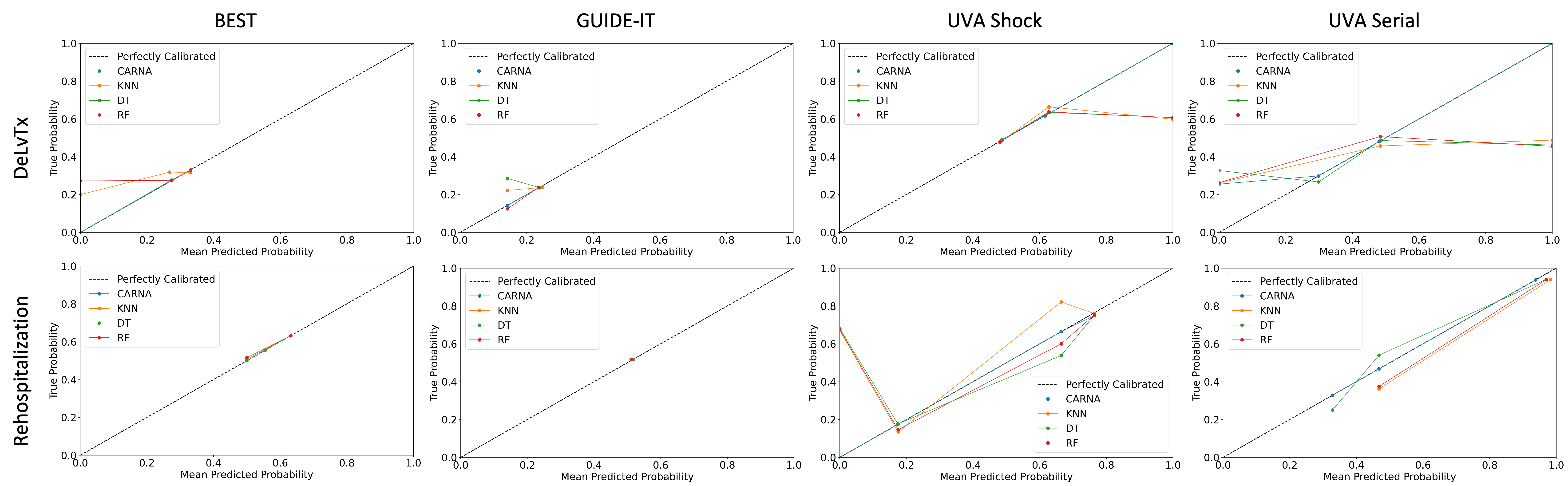}
    \caption{Calibration Plots for All Features using a bin size of 10. True probability is the fraction of positives per bin.}
    \label{fig:calib-allfts}
\end{figure*}

\begin{figure}
    \centering
    \includegraphics[width=\linewidth]{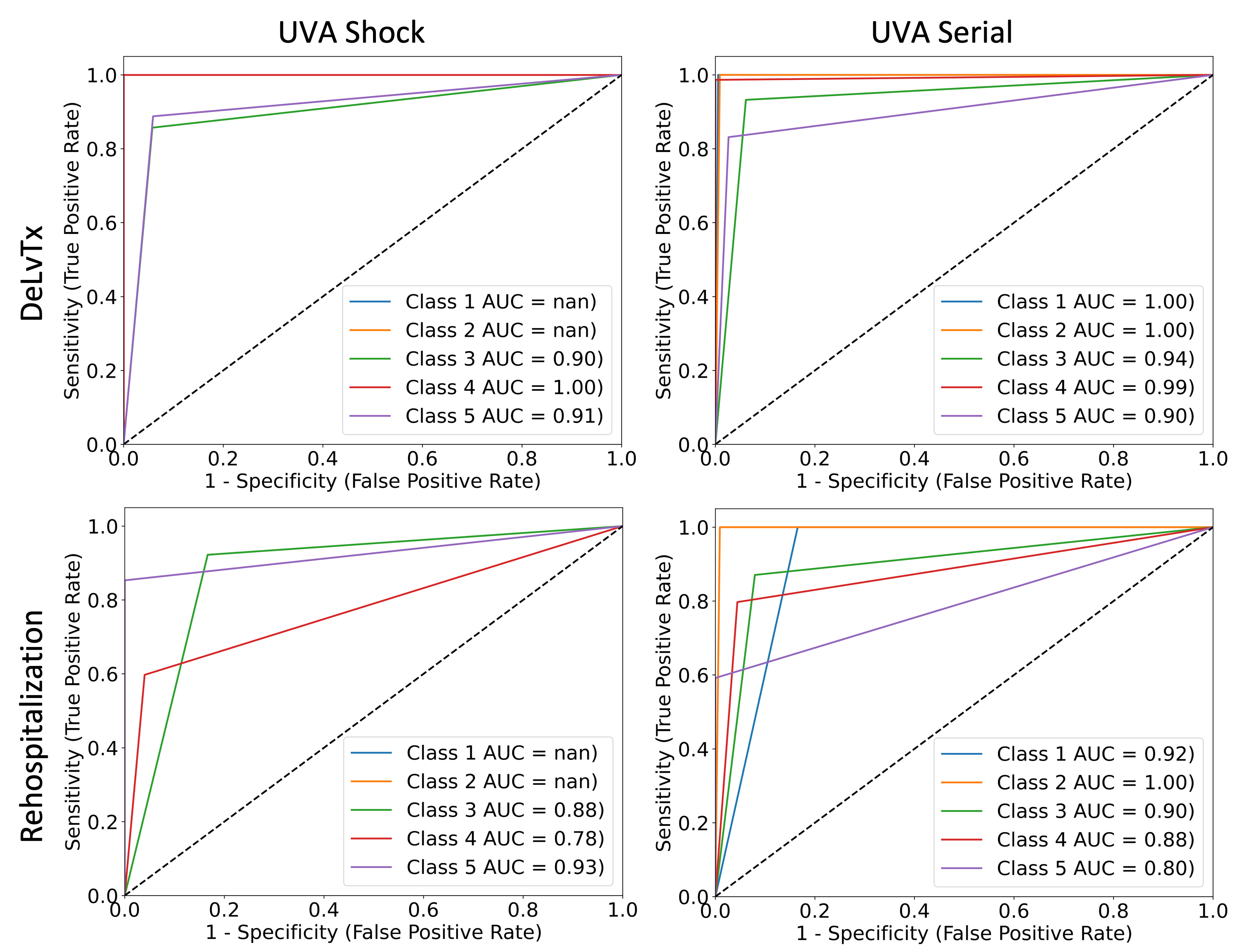}
    \caption{ROC Curves for Validation Datasets and Invasive Hemodynamics Feature Set. nan = no data in that class.}
    \label{fig:carna-hemo-roc}
\end{figure}

\begin{figure}[t]
    \centering
    \includegraphics[width=\linewidth]{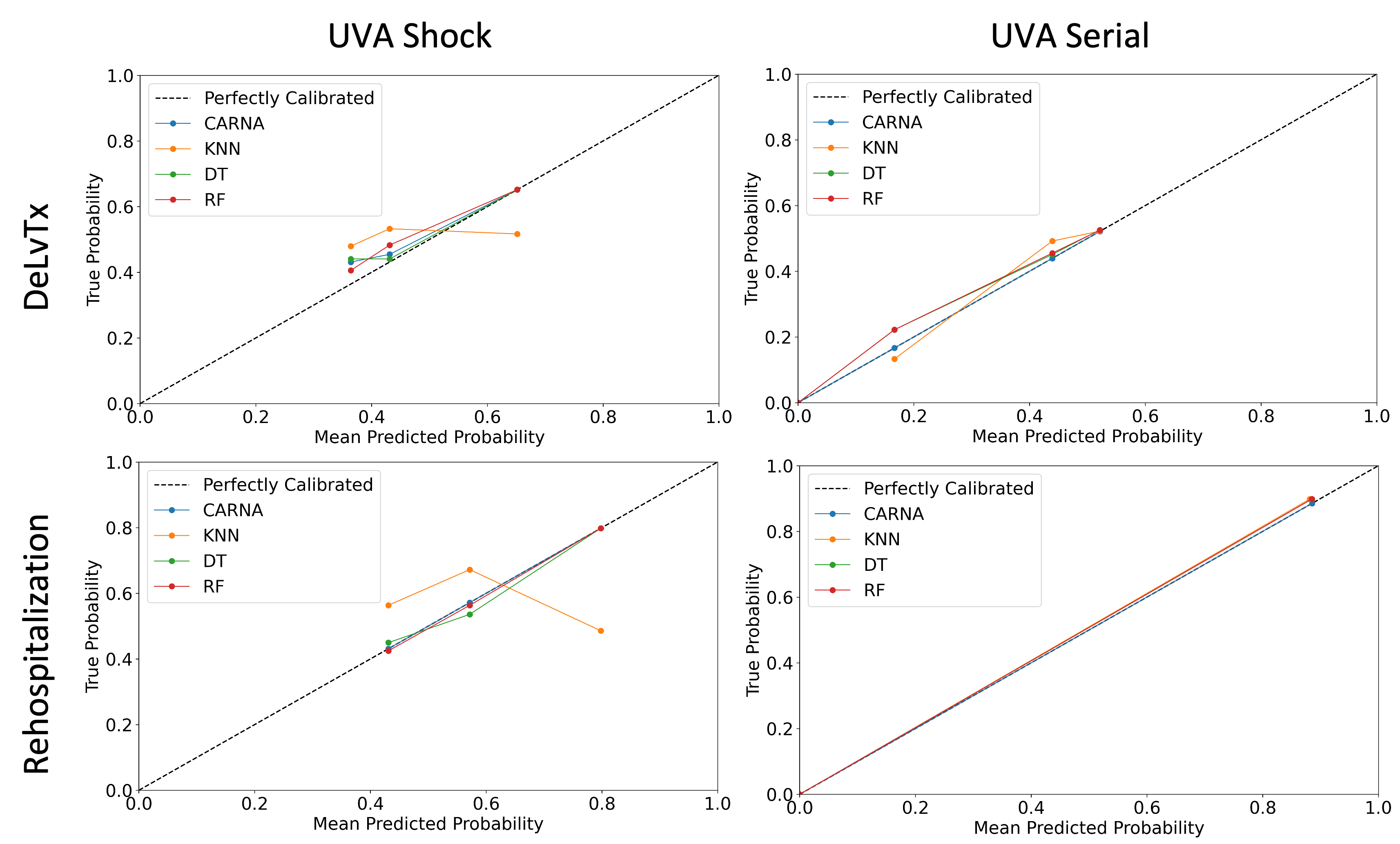}
    \caption{Calibration Plots for Invasive Hemodynamics using bin size of 10. True probability is the fraction of positives per bin.} 
    % We note that some bins have no samples, hence why some plots do not have complete points in the line graphs.}
    \label{fig:calib-hemo}
\end{figure}

\begin{table}
\caption{CARNA Comparison to Traditional ML Models - Invasive Hemodynamics Feature Set}\label{table:carna-trad-ml-ih}
\centering
\caption*{DeLvTx Outcome}
\resizebox{\linewidth}{!}{%
\begin{tabular}{cccccc} \hline \toprule
\textbf{Dataset} & \textbf{Model} & \textbf{Accuracy} & \textbf{Averaged AUC} & \textbf{Sensitivity} & \textbf{Specificity} \\ \midrule
\multirow{4}{*}{UVA Shock} & CARNA & \textbf{0.947$\pm$0.107}	& \textbf{0.938$\pm$0.106}	& \textbf{0.915$\pm$0.103}	& \textbf{0.961$\pm$0.108} \\
 & KNN & 0.660$\pm$0.064	& 0.471$\pm$0.027	& 0.154$\pm$0.094	& 0.789$\pm$0.086 \\ 
 & DT & 0.759$\pm$0.081	& 0.628$\pm$0.057	& 0.405$\pm$0.049	& 0.85$\pm$0.094 \\
 & RF & 0.754$\pm$0.08	& 0.637$\pm$0.059	& 0.426$\pm$0.043	& 0.849$\pm$0.094 \\ \midrule
\multirow{4}{*}{UVA Serial} & CARNA &  \textbf{0.969$\pm$0.081}	& \textbf{0.965$\pm$0.080}	& \textbf{0.950$\pm$0.079}	& \textbf{0.980$\pm$0.081} \\
 & KNN & 0.685$\pm$0.051	& 0.5$\pm$0.001	& 0.199$\pm$0.065	& 0.801$\pm$0.065 \\ 
 & DT & 0.776$\pm$0.062	& 0.649$\pm$0.045	& 0.438$\pm$0.029	& 0.86$\pm$0.071 \\
 & RF & 0.772$\pm$0.061	& 0.628$\pm$0.042	& 0.4$\pm$0.037	& 0.855$\pm$0.07 \\ \bottomrule
\end{tabular}
}
\vspace*{\baselineskip}
\caption*{Rehospitalization Outcome}
\resizebox{\linewidth}{!}{%
\begin{tabular}{cccccc} \hline \toprule
\textbf{Data Set} & \textbf{Model} & \textbf{Accuracy} & \textbf{Averaged AUC} & \textbf{Sensitivity} & \textbf{Specificity} \\ \midrule
\multirow{4}{*}{UVA Shock} & CARNA & \textbf{0.907$\pm$0.102}	& \textbf{0.861$\pm$0.096} & \textbf{0.791$\pm$0.086} & \textbf{0.935$\pm$0.105} \\
 & KNN & 0.66$\pm$0.064	& 0.471$\pm$0.027	& 0.154$\pm$0.094	& 0.789$\pm$0.086 \\
 & DT & 0.759$\pm$0.081	& 0.628$\pm$0.057	& 0.405$\pm$0.049	& 0.85$\pm$0.094 \\
 & RF & 0.754$\pm$0.08	& 0.637$\pm$0.059	& 0.426$\pm$0.043	& 0.849$\pm$0.094 \\ \midrule
\multirow{4}{*}{UVA Serial} & CARNA &  \textbf{0.896$\pm$0.074}	& \textbf{0.896$\pm$0.074}	& \textbf{0.852$\pm$0.070}	& \textbf{0.940$\pm$0.078} \\
 & KNN & 0.685$\pm$0.051	& 0.5$\pm$0.001	& 0.199$\pm$0.065	& 0.801$\pm$0.065\\ 
 & DT & 0.776$\pm$0.062	& 0.649$\pm$0.045	& 0.438$\pm$0.029 &	0.86$\pm$0.071 \\
 & RF & 0.772$\pm$0.061	& 0.628$\pm$0.042	& 0.4$\pm$0.037	& 0.855$\pm$0.07 \\ \bottomrule
\end{tabular}
}
\parbox{\linewidth}{\caption*{\scriptsize Table reports value$\pm$confidence interval, bolded values indicate highest scoring item in each block. KNN = K-Nearest Neighbor; DT = Decision Tree, RF = Random Forest; DeLvTx = composite endpoint of death, LVAD implantation or transplantation.}}
\end{table}

\begin{table}[t]
\caption{CARNA Comparison to Traditional ML Models - All Features Feature Set}\label{table:carna-trad-ml-allfts}
\centering
\caption*{DeLvTx Outcome}
\resizebox{\linewidth}{!}{%
\begin{tabular}{cccccc} \hline \toprule
\textbf{Data Set} & \textbf{Model} & \textbf{Accuracy} & \textbf{Averaged AUC} & \textbf{Sensitivity} & \textbf{Specificity} \\ \midrule
\multirow{4}{*}{BEST} & CARNA & \textbf{0.997$\pm$0.037}	& \textbf{0.994$\pm$0.037}	& \textbf{0.990$\pm$0.037}	& \textbf{0.998$\pm$0.037} \\
 & KNN & 0.822$\pm$0.03	& 0.677$\pm$0.022	& 0.525$\pm$0.008	& 0.83$\pm$0.03 \\
 & DT & 0.831$\pm$0.03	& 0.652$\pm$0.021	& 0.469$\pm$0.009	& 0.835$\pm$0.03 \\
 & RF & 0.844$\pm$0.031	& 0.722$\pm$0.025	& 0.602$\pm$0.017	& 0.842$\pm$0.031 \\ \midrule
\multirow{4}{*}{GUIDE-IT} & CARNA & \textbf{0.997$\pm$0.070}	& \textbf{0.996$\pm$0.070}	& \textbf{0.995$\pm$0.069}	& \textbf{0.998$\pm$0.070} \\
 & KNN & 0.849$\pm$0.058 & 0.575$\pm$0.027	& 0.295$\pm$0.045	& 0.854$\pm$0.059 \\
 & DT & 0.852$\pm$0.058	& 0.572$\pm$0.026	& 0.277$\pm$0.047	& 0.866$\pm$0.06 \\
 & RF & 0.955$\pm$0.067	& 0.719$\pm$0.046	& 0.475$\pm$0.016	& 0.964$\pm$0.067 \\ \midrule
\multirow{4}{*}{UVA Shock} & CARNA & \textbf{0.865$\pm$0.049}	& \textbf{0.871$\pm$0.050} & \textbf{0.811$\pm$0.045}	& \textbf{0.931$\pm$0.054} \\
 & KNN & 0.349$\pm$0.032	& 0.391$\pm$0.027	& 0.042$\pm$0.055	& 0.624$\pm$0.029 \\
 & DT & 0.386$\pm$0.028	& 0.319$\pm$0.035	& 0.135$\pm$0.049	& 0.596$\pm$0.025 \\
 & RF & 0.604$\pm$0.026	& 0.377$\pm$0.029	& 0.016$\pm$0.057	& 0.771$\pm$0.042 \\ \midrule
\multirow{4}{*}{UVA Serial} & CARNA & \textbf{0.858$\pm$0.067}	& \textbf{0.871$\pm$0.068}	& \textbf{0.815$\pm$0.063}	& \textbf{0.927$\pm$0.073} \\
 & KNN & 0.346$\pm$0.044	& 0.409$\pm$0.034	& 0.022$\pm$0.077	& 0.612$\pm$0.037 \\
 & DT & 0.459$\pm$0.023	& 0.383$\pm$0.038	& 0.208$\pm$0.061	& 0.596$\pm$0.035 \\
 & RF & 0.339$\pm$0.045	& 0.4$\pm$0.035	& 0.009$\pm$0.079	& 0.606$\pm$0.037 \\ \bottomrule
\end{tabular}
}
\vspace*{\baselineskip}
\caption*{Rehospitalization Outcome}
\resizebox{\linewidth}{!}{%
\begin{tabular}{cccccc} \hline \toprule
\textbf{Data Set} & \textbf{Model} & \textbf{Accuracy} & \textbf{Averaged AUC} & \textbf{Sensitivity} & \textbf{Specificity} \\ \midrule
\multirow{4}{*}{BEST} & CARNA & \textbf{0.997$\pm$0.037}	& \textbf{0.994$\pm$0.037}	& \textbf{0.990$\pm$0.037}	& \textbf{0.998$\pm$0.037} \\
 & KNN & 0.822$\pm$0.03	& 0.677$\pm$0.022	& 0.525$\pm$0.008	& 0.83$\pm$0.03 \\
 & DT & 0.831$\pm$0.03	& 0.652$\pm$0.021	& 0.469$\pm$0.009	& 0.835$\pm$0.03 \\
 & RF & 0.844$\pm$0.031	& 0.722$\pm$0.025	& 0.602$\pm$0.017	& 0.842$\pm$0.031 \\ \midrule
\multirow{4}{*}{GUIDE-IT} & CARNA & \textbf{0.997$\pm$0.070}	& \textbf{0.996$\pm$0.070}	& \textbf{0.995$\pm$0.069}	& \textbf{0.998$\pm$0.070} \\
 & KNN & 0.611$\pm$0.033	& 0.646$\pm$0.038	& 0.489$\pm$0.01	& 0.803$\pm$0.054 \\
 & DT & 0.612$\pm$0.033	& 0.582$\pm$0.028	& 0.369$\pm$0.036	& 0.794$\pm$0.053 \\
 & RF & 0.611$\pm$0.033	& 0.623$\pm$0.035	& 0.446$\pm$0.023	& 0.801$\pm$0.054 \\ \midrule
\multirow{4}{*}{UVA Shock} & CARNA & \textbf{0.694$\pm$0.036}	& \textbf{0.533$\pm$0.015} & \textbf{0.250$\pm$0.041}	& \textbf{0.816$\pm$0.046} \\ 
 & KNN & 0.349$\pm$0.032	& 0.391$\pm$0.027	& 0.042$\pm$0.055	& 0.624$\pm$0.029 \\ 
 & DT & 0.386$\pm$0.028	& 0.319$\pm$0.035	& 0.135$\pm$0.049	& 0.596$\pm$0.025 \\
 & RF & 0.604$\pm$0.026	& 0.377$\pm$0.029	& 0.016$\pm$0.057	& 0.771$\pm$0.042 \\ \midrule
\multirow{4}{*}{UVA Serial} & CARNA & \textbf{0.890$\pm$0.070}	& \textbf{0.798$\pm$0.061}	& \textbf{0.653$\pm$0.044}	& \textbf{0.942$\pm$0.075} \\
 & KNN & 0.346$\pm$0.044	& 0.409$\pm$0.034	& 0.022$\pm$0.077	& 0.612$\pm$0.037 \\
 & DT & 0.459$\pm$0.023	& 0.383$\pm$0.038	& 0.208$\pm$0.061	& 0.596$\pm$0.035 \\
 & RF & 0.339$\pm$0.045	& 0.4$\pm$0.035	& 0.009$\pm$0.079	& 0.606$\pm$0.037 \\ \bottomrule
\end{tabular}
}
\parbox{\linewidth}{\caption*{\scriptsize Table reports value$\pm$confidence interval, bolded values indicate highest scoring item in each block. KNN = K-Nearest Neighbor; DT = Decision Tree, RF = Random Forest; DeLvTx = composite endpoint of death, LVAD implantation or transplantation.}}
\end{table}

\subsection{Risk Label Generation Results}
From the elbow plots, 5 was chosen as the optimal number of cluster groups corresponding to 5 risk categories.  
% We selected five as the optimal number of groups based on the elbow plots corresponding to 5 risk categories; 
% We selected 5 clustering groups corresponding to 5 risk categories for all models based on the elbow plots. 
% , shown in Supplemental Figure S1. 
An example dendrogram displaying the cluster splits for the All Features feature set is shown in Figure~\ref{fig:dendograms}. In hierarchical clustering methods, clusters are separated by horizontally dividing the top of the hierarchy based on the specified number of groups, illustrated by the horizontal dashed black line in the figures. 
% (5 in our case). This is illustrated by the horizontal dashed black line in the figures. 
% Each leaf (end of the dendrogram) represents an individual data point. 
% Our clusters are distinct and have a high degree of separation, with low C Indexes of 0.063 for the Hemodynamics feature set and 0.051 for the All Data feature set.
\textbf{Our clusters are distinct with a high degree of separation, with low C Indexes of 0.063 for the Hemodynamics feature set and 0.051 for the All Features feature set.}

Table~\ref{table:risk-cats} reports the risk score meaning and corresponding real-valued average risk probabilities for each score category across all feature sets and outcomes. For example, for the Hemodynamics feature set and the DeLvTx outcome, a risk score of 3 indicates a 20–30\% chance of the outcome, with a mean outcome probability of 0.245 computed from the patients in this cluster. For a sanity check, we also reported the average risk probabilities for each dataset individually. \textbf{These results provide evidence that the risk ranges correspond to the real observed risk in the patient cohorts.}

\subsection{Learned MVDDs}
We generated a total of four MVDD models for each of the feature sets (Invasive Hemodynamics and All Features) and outcomes (DeLvTx and Rehospitalization). 
% Figure~\ref{fig:example-mvdd} shows an example of a subset of a learned MVDD for the Invasive Hemodynamics feature set and the outcome DeLvTx. 
The Invasive Hemodynamic models use a combination of 28 features that include basic demographics, invasive and noninvasive hemodynamics; the All Features models use a combination of 66 features across demographics, labs, medications, exercise, quality metrics, other medical diagnostics and noninvasive hemodynamics. We note that these are the \textit{maximum} number of features per model and actual prediction paths through the MVDDs use smaller subsets with interchangeable combinations of features (e.g., the features that may be “or-ed” together along a path that provide choices for which feature is used for prediction in the phenotype.) 
% A complete list of the features used in the models is available in Supplemental Table S2. 
% Figures of the full MVDD models are available in Supplemental Figure S4, and a discussion of the identified phenotypes is available in the Supplemental Results.

% \subsection{Performance of the Models}
\subsubsection{MVDD Performance} Table~\ref{table:carna-performance} presents the validation performance summary. The UVA Cardiogenic Shock and Serial Cardiac cohorts were used to validate the invasive hemodynamics models, since they were the only cohorts with invasive hemodynamics; all 4 validation cohorts were used to validate the All Features models. Figures \ref{fig:carna-allfts-roc} and \ref{fig:carna-hemo-roc} show the ROC curves and AUC values for each risk class for the All Features and Invasive Hemodynamics sets, respectively. For the Invasive Hemodynamics feature set across all outcomes, our validation models performed well with averaged AUCs of 0.861$\pm$0.096 to 0.965$\pm$0.080. For the All Features set across all datasets, our models performed well for the DeLvTx outcome with averaged AUCs of 0.871$\pm$0.068 to 0.996$\pm$0.070 and moderately for the rehospitalization outcome with averaged AUCs of 0.533$\pm$0.015 to 0.996$\pm$0.070. \textbf{These validation results provide evidence that the CARNA models yield robust risk stratification.}

\subsection{Comparison to Traditional ML Models}
For additional comparison, the performance of the CARNA models was compared with traditional ML models, including K-Nearest Neighbors (KNN), Decision Trees (DT) and Random Forests (RF), reported in Table~\ref{table:carna-trad-ml-ih} and \ref{table:carna-trad-ml-allfts} for the Invasive Hemodynamic and All Features feature sets, respectively. 
Calibration plots are also shown in Figures~\ref{fig:calib-allfts} and \ref{fig:calib-hemo}. Some bins have no samples, hence why some plots do not have complete points in the line graphs. Of the traditional models, RFs followed by DTs tend to perform the best. \textbf{Across all feature sets, outcomes and datasets, the CARNA models outperform traditional ML models}.

% \subsection{Identified Phenotypes}
% In the MVDDs, the features towards the top of the graph are the most predictive. Across the Invasive Hemodynamics feature set, sex, SBP, CPI, MAP, Pulmonary Artery Pulsatility Index (PAPi), PAS, MPAP, Pulmonary Arterial Proportional Pulse Pressure (Right Side, PAPP), the ratio of RAP to PCWP (RAT), and the ratio of systemic pulse pressure to HR (PPratio) were consistently some of the most predictive features. Across the All Features feature set, weight, sex, age, SBP, DBP, HR, MAP, Systemic Pulse Pressure (PP), Systemic Arterial Proportional Pulse Pressure (PPP), and lab values including creatinine, potassium, blood urea nitrogen, sodium and white blood cell count were consistently some of the most predictive features. 
% Supplemental Figure S6 displays example phenotypes from the paths in the learned MVDDs for the Invasive Hemodynamics and All Features feature sets.

% \begin{figure}
%     \centering
%     \includegraphics[width=\linewidth]{Figures/web_portal_invasive.png}
%     \caption{Example CARNA Web Portal -- interface for predicting the invasive hemodynamic risk score. }
%     \label{fig:web-portal}
% \end{figure}

\subsection{Comparison to Previous HF Risk Scores}
Benchmark comparison between the CARNA risk scores and 6 other previously developed HF risk scores are shown in Table~\ref{table:carna-score-comparison}. The table reports the AUCs for each of the HF risk scores on all datasets. Table~\ref{table:carna-hypoth-testing} displays results of the hypothesis testing between CARNA and previous scores. The delta AUC and p-values are reported; a p-value of $<$0.05 indicates there is a significant difference between the two scores. \textbf{CARNA outperforms all previous HF risk scores.}

\begin{table*}[t]
\caption{Comparison to Previous Scores - AUC for Outcome Mortality}\label{table:carna-score-comparison}
\centering
\resizebox{0.9\linewidth}{!}{%
\begin{tabular}{c|c|ccccc} \hline \toprule
\multirow{2}{*}{\textbf{Score}} & \textbf{Median} & \multicolumn{5}{c}{\textbf{Dataset}} \\ 
 & \textbf{Follow-Up} & ESCAPE & BEST & GUIDE-IT & UVA Shock & UVA Serial \\ \midrule
CARNA - Hemo & 6 months	& 0.952$\pm$0.091	& N/A	& N/A	& \textbf{0.938$\pm$0.106}	& \textbf{0.965$\pm$0.080} \\
CARNA - All Fts	& 6 months	& \textbf{0.978$\pm$0.065} & \textbf{0.994$\pm$0.037}	& \textbf{0.996$\pm$0.070}	& 0.871$\pm$0.050	& 0.871$\pm$0.068 \\
ADHERE~\cite{fonarow2005risk} & 5.85 days	& 0.595$\pm$0.029	& 0.576$\pm$0.015	& 0.601$\pm$0.021	& 0.526$\pm$0.013	& 0.574$\pm$0.030 \\
EFFECT 30D~\cite{lee2003predicting} & 30 days	& 0.550$\pm$0.021	& 0.610$\pm$0.018 &	0.635$\pm$0.024	& 0.584$\pm$0.024	& 0.610$\pm$0.037 \\
EFFECT Y1~\cite{lee2003predicting} & 1 year	& 0.548$\pm$0.021	& 0.638$\pm$0.020	& 0.632$\pm$0.024	& 0.612$\pm$0.027	& 0.644$\pm$0.043 \\
ESCAPE~\cite{o2010escapescore} &	6 months	& 0.681$\pm$0.057	& 0.587$\pm$0.016	& 0.715$\pm$0.043	& 0.595$\pm$0.025	& 0.565$\pm$0.029 \\ 
GWTG~\cite{peterson2010validated} & 4 days	& 0.601$\pm$0.030	& 0.538$\pm$0.010	& 0.537$\pm$0.013	& N/A	& N/A \\ 
MAGGIC Y1~\cite{pocock2013predicting} & 2.5 years	& 0.640$\pm$0.035	& N/A	& 0.689$\pm$0.029	& 0.678$\pm$0.034	& N/A \\ 
MAGGIC Y3~\cite{pocock2013predicting} & 2.5 years	& 0.640$\pm$0.035	& N/A	& 0.689$\pm$0.029	& 0.678$\pm$0.034	& N/A \\ 
SHFM Y1~\cite{levy2006seattle} & 1 year	& 0.623$\pm$0.033	& 0.613$\pm$0.018	& 0.623$\pm$0.023	& 0.587$\pm$0.024	& 0.588$\pm$0.033 \\
SHFM Y3~\cite{levy2006seattle} & 3 years	& 0.623$\pm$0.033	& 0.616$\pm$0.018	& 0.625$\pm$0.023	& 0.588$\pm$0.024	& 0.584$\pm$0.032 \\ 
SHFM Y5~\cite{levy2006seattle} & 5 years	& 0.622$\pm$0.033	& 0.615$\pm$0.018	& 0.619$\pm$0.023	& 0.573$\pm$0.022	& 0.579$\pm$0.032 \\ \bottomrule
\end{tabular}
}
\parbox{0.9\linewidth}{\caption*{\scriptsize Table displays AUC$\pm$confidence interval; bolded values indicate highest performing score for each dataset. N/A = score could not be calculated for the dataset; Hemo = Invasive Hemodynamic; All Fts = All Features; 30D = 30-day mortality; Y1 = 1 year mortality; Y3 = 3 year mortality; Y5 = 5 year mortality.}}
\end{table*}

\begin{table*}
\caption{Hypothesis Testing Between CARNA and Comparison HF Scores}\label{table:carna-hypoth-testing}
\centering
\resizebox{\linewidth}{!}{%
\begin{tabular}{c|ccc|ccccc} \hline \toprule
& \multicolumn{3}{|c|}{\textbf{Invasive Hemodynamic Feature Set}} & \multicolumn{5}{|c}{\textbf{All Features Feature Set}} \\ \midrule
\multirow{2}{*}{\textbf{Score}}  & \multicolumn{3}{|c|}{\textbf{Dataset}} & \multicolumn{5}{|c}{\textbf{Dataset}} \\ 
 & ESCAPE  & UVA Shock & UVA Serial & ESCAPE  & BEST & GUIDE-IT  & UVA Shock & UVA   Serial\\ \midrule 
ADHERE~\cite{fonarow2005risk} & -0.357,   0.262 & -0.412,   \textless{}0.001 & -0.391,   0.413 & -0.383,   0.031 & -0.418,   \textless{}0.001 & -0.395,   0.005 & -0.345,   \textless{}0.001 & -0.297,   0.413 \\
EFFECT   30D~\cite{lee2003predicting} & -0.402,   0.881 & -0.354,   \textless{}0.001 & -0.355,   0.315 & -0.163,   0.020 & -0.428,   \textless{}0.001 & -0.384,   0.069 & -0.361,   \textless{}0.001 & -0.287,   0.315 \\
EFFECT   Y1~\cite{lee2003predicting} & -0.404,   0.832 & -0.326,   \textless{}0.001 & -0.321,   0.028 & -0.430,   0.026 & -0.356,   \textless{}0.001 & -0.364,   0.096 & -0.259,   \textless{}0.001 & -0.227,   0.028 \\
ESCAPE~\cite{o2010escapescore} & -0.271,   0.008 & -0.343,   \textless{}0.001 & -0.400,   0.311 & -0.297,   \textless{}0.001 & -0.407,   \textless{}0.001 & -0.281,   \textless{}0.001 & -0.276,   \textless{}0.001 & -0.306,   0.311 \\
GWTG~\cite{peterson2010validated}& -0.351,   0.593 & N/A & N/A & -0.377,   0.002 & -0.456,   0.001 & -0.459,   0.001 & N/A & N/A \\
MAGGIC   Y1~\cite{pocock2013predicting} & -0.312,   0.151 & -0.260,   0.018 & N/A & -0.338,   0.094 & N/A & -0.307,   0.048 & -0.193,   0.018 & N/A \\
MAGGIC   Y3~\cite{pocock2013predicting} & -0.312,   0.151 & -0.260,   0.018 & N/A & -0.338,   0.094 & N/A & -0.307,   0.048 & -0.193,   0.018 & N/A \\
SHFM   Y1~\cite{levy2006seattle} & -0.329,   0.011 & -0.351,   \textless{}0.001 & -0.377,   0.784 & -0.355,   \textless{}0.001 & -0.381,   \textless{}0.001 & -0.373,   0.201 & -0.284,   \textless{}0.001 & -0.283,   0.784 \\
SHFM   Y3~\cite{levy2006seattle} & -0.329,   0.012 & -0.350,   \textless{}0.001 & -0.381,   0.878 & -0.355,   \textless{}0.001 & -0.378,   \textless{}0.001 & -0.371,   0.173 & -0.283,   \textless{}0.001 & -0.287,   0.878 \\
SHFM   Y5~\cite{levy2006seattle} & -0.330,   0.013 & -0.365,   \textless{}0.001 & -0.386,   0.996 & -0.356,   \textless{}0.001 & -0.379,   \textless{}0.001 & -0.377,   0.268 & -0.298,   \textless{}0.001 & -0.292,   0.996 \\ \bottomrule
\end{tabular}
}
\caption*{\scriptsize Table reports $\Delta$AUC, p-value. N/A = score could not be calculated for the dataset; 30D = 30-day mortality; Y1 = 1 year mortality; Y3 = 3 year mortality; Y5 = 5 year mortality.}
\end{table*}
\section{Discussion}
% New Discussion Outline
% Summary of approach 
% Discussion of results - robust risk stratification
% HF specific findings
% Limitations & approach choices
% Use of framework for other purposes, and clinical risk strat and phenotyping, how can be used, 

In this paper, we developed an explainable ML approach using Multi-Valued Decision Diagrams to derive and validate a novel HF risk score that incorporates invasive hemodynamic and other clinical variables to stratify risk of adverse outcomes in advanced HF patients. The CARNA risk scores were highly predictive of adverse outcomes in a broad spectrum of HF patients. Accurately identifying high-risk advanced HF patients early on is fundamental for timely allocation of life-saving therapies and improvement of patient outcomes. 

\subsubsection{Study Strengths} This study has several strengths relative to previous approaches. First, our models use a richer, more diverse feature set, beyond what is used in many other clinical risk scores. Second, the use of both invasive and noninvasive hemodynamics provides incremental utility in outcome prediction, as evidenced by robust validation metrics. Third, we can make predictive decisions using incomplete data, which is particularly advantageous in clinical scenarios with missing data. Fourth, we use multiple patient cohorts to train and validate our models, compare risk ascertainment to previous HF risk scores, and compare our models with traditional ML models, which support CARNA’s application to diverse real-world patient cohorts. Finally, our ML method is explainable and interpretable; elucidation of the phenotypes used to make risk characterizations by our models allow clinicians to better understand how and why a risk score was given. These phenotypes may identify possible HF subgroups that can be further investigated in clinical studies.

\subsubsection{CARNA Outperforms Benchmarks} As shown in Tables~\ref{table:carna-performance}--\ref{table:carna-hypoth-testing}, the CARNA risk scores highly outperform previous risk scores. The CARNA Invasive Hemodynamics score was more predictive than other scores including the ESCAPE risk score which was derived on the same cohort as our training data using linear statistical methods. The CARNA All Features score also outperformed previous risk scores, with the exception of the Rehospitalization outcome for the two UVA cohorts, which performed similar to standard risk models. Moreover, as evidenced by Tables \ref{table:carna-trad-ml-ih} and \ref{table:carna-trad-ml-allfts}, the CARNA models outperform traditional ML models across all datasets, feature sets and outcomes. We speculate MVDDs may outperform traditional ML models due to their ability to handle missing data.

\subsubsection{Comment on Hemodynamics} The CARNA Invasive Hemodynamic models do better than the CARNA All Features models, which suggests that invasive hemodynamics (along with integrated metrices) improve outcome prediction for advanced HF patients. Integrated hemodynamic indices such as Cardiac Power Index, Mean Arterial Pressure, and Pulmonary Artery Pulsatility Index were highly predictive of patient outcomes. This aligns with findings from previous studies, demonstrating the incremental utility of integrated metrics in risk assessment~\cite{bilchick2018clinical,bilchick2018plasma,mazimba2016decreased,mazimba2022systemic}.

\subsubsection{Model Design Choices and Limitations}
Our models use single point-of-care measurements, and do not take advantage of multiple follow-up recordings. As a result, they may lose interrelations available from multiple temporal recordings (i.e., changes between measurements). However, using single measurements in our models allows for clinician ease-of-use. 
Furthermore, although only the “OR” nodes in the MVDD model explicitly handle missing data, we chose to use ``AND/OR” MVDDs because the ``OR”-only MVDDs become very large and overfit the data. We used single MVDD models for interpretability purposes throughout the project evaluation. However, ensemble approaches (e.g., ensembles of MVDDs) have been shown to outperform single model methods~\cite{hosni2021systematic}, and this will be investigated in future.
Additionally, we note that an aspect of model interpretability may be lost due to the model predicting risk classes generated from an unsupervised clustering method as opposed to predicting the binary outcome(s) directly. Even so, we believe such a tradeoff may be acceptable due to the improved ability to risk-stratify HF patients. 

Despite fewer patients and shorter follow-up time (6-months) compared to other datasets, the ESCAPE trial was selected for model training because, to the best of the authors’ knowledge, it is the only cohort available with detailed invasive hemodynamics derived from a well-designed randomized HF clinical trial. 
There is potential for selection bias by choosing trial data and higher-risk patients in the two UVA cohorts. In addition, many of our validation datasets did not have invasive hemodynamics so we were unable to validate the invasive hemodynamic models on all four of the patient cohorts.  Further, there were heterogeneities in HF acuity status in the datasets used. Even so, validation of the CARNA models yielded robust risk stratification compared to other conventional HF risk score and ML models.

\section{Conclusion}
This study developed a novel advanced HF risk stratification using an explainable ML methodology. The CARNA risk scores are more predictive of patient outcomes than previous approaches and provide detailed characterizations of clinical phenotypes. CARNA may facilitate clinical decision making and provide robust risk stratification.

% \section{Translational Outlook}
% In order to facilitate clinical use and promote open science, we have developed a tool implementation, entitled CARNA (Characterizing Advanced heart failure Risk and hemodyNAmic phenotypes) so named for the roman healing goddess who presides over the heart, available here: https://github.com/jozieLamp/CARNA. Our implementation includes a web server (Supplemental Figure S7), which provides live CARNA risk score prediction here: http://hemopheno.pythonanywhere.com/.

% \appendices
% Appendixes, if needed, appear before the acknowledgment.

% \section*{Acknowledgment}
% The preferred spelling of the word ``acknowledgment'' in American English is 
% without an ``e'' after the ``g.'' Use the singular heading even if you have 
% many acknowledgments. Avoid expressions such as ``One of us (S.B.A.) would 
% like to thank $\ldots$ .'' Instead, write ``F. A. Author thanks $\ldots$ .'' In most 
% cases, sponsor and financial support acknowledgments are placed in the 
% unnumbered footnote on the first page, not here.

% \section*{REFERENCES}
\printbibliography

\end{document}